\documentclass[11pt]{article}

\usepackage[final]{acl}
\usepackage{amsmath}
\usepackage{booktabs}
\usepackage{array}
\usepackage{amsmath}
\usepackage{amssymb}
\usepackage{times}
\usepackage{latexsym}
\usepackage{times}
\usepackage{latexsym}
\usepackage{amsmath}
\usepackage{amsfonts}
\usepackage{amssymb}
\usepackage{algorithm}
\usepackage{xcolor}
\usepackage{algpseudocode}
\usepackage{csquotes}   
\usepackage{graphicx}
\usepackage{stfloats}
\usepackage{amsmath}
\usepackage{amssymb}
\usepackage{graphicx}
\usepackage{hyperref}
\usepackage{amsthm}
\usepackage{multirow}
\newtheorem{definition}{Definition}
\usepackage{graphicx} 
\usepackage[utf8]{inputenc} 

\usepackage{pifont} 

\usepackage[T1]{fontenc}

\usepackage[utf8]{inputenc}

\usepackage{microtype}

\usepackage{inconsolata}

\usepackage{graphicx}
\usepackage{booktabs} 
\usepackage{siunitx} %
\usepackage{tikz}
\usetikzlibrary{arrows.meta}
\usepackage{graphicx}
\usepackage{tikz}
\usetikzlibrary{arrows.meta}

\title{CMR Scaling Law: Predicting Critical Mixture Ratios for Continual Pre-training of Language Models}

\author{
Jiawei Gu$^{\S,\dag,1,2}$, Zacc Yang$^{\dag,2}$, Chuanghao Ding$^{2,3}$, Rui Zhao$^2$, Fei Tan$^{*,2}$ \\
$^1$Sun Yat-sen University\\ $^2$SenseTime Research \\
$^3$Nanjing University \\
\small
\tt $^{1}$kuvvius@gmail.com\ \
\small
\tt $^{2}$\{yangzacc, zhaorui, tanfei\}@sensetime.com\ \
\small
\tt $^{3}$ch777.ding@smail.nju.edu.cn
}

\begin{document}
\maketitle
\def\thefootnote{$\S$}\footnotetext{Work was done during internship at SenseTime Research}
\def\thefootnote{$\dag$}\footnotetext{Equal contribution}
\def\thefootnote{*}\footnotetext{Corresponding author}
\begin{abstract}

Large Language Models (LLMs) excel in diverse tasks but often underperform in specialized fields due to limited domain-specific or proprietary corpus. Continual pre-training (CPT) enhances LLM capabilities by imbuing new domain-specific or proprietary knowledge while replaying general corpus to prevent catastrophic forgetting. The data mixture ratio of general corpus and domain-specific corpus, however, has been chosen heuristically, leading to sub-optimal training efficiency in practice. 
In this context, we attempt to re-visit the scaling behavior of LLMs under the hood of CPT, and discover a power-law relationship between loss, mixture ratio, and training tokens scale. 
We formalize the trade-off between general and domain-specific capabilities, leading to a well-defined Critical Mixture Ratio (CMR) of general and domain data. 
By striking the balance, CMR maintains the model's general ability and achieves the desired domain transfer, ensuring the highest utilization of available resources. Considering the balance between efficiency and effectiveness, CMR can be regarded as the optimal mixture ratio. Through extensive experiments, we ascertain the predictability of CMR, propose CMR scaling law and have substantiated its generalization. These findings offer practical guidelines for optimizing LLM training in specialized domains, ensuring both general and domain-specific performance while efficiently managing training resources.

\end{abstract}

\section{Introduction}
    
Large Language Models (LLMs) exhibit versatile abilities, including question answering, translation, summarization, role-playing, etc.~\cite{gpt3,llama,llama2,starcoder,lu-makes-plm-0-shot}. Their performance, however, may degrade in specific domains due to limited corresponding pre-training data. To enhance LLMs' abilities in specialized areas and avoid the enormous cost of re-training, a popular approach is Continual Pre-Training (CPT)~\cite{law_llm,med_llm,cpt_llm_survey,empirical_cv_catastrophic_forgetting}. This approaches are likely to equip LLMs with new domain-related capabilities without much general performance penalty.

Although CPT has been proven effective on multiple domains such as code~\cite{starcoder,autocoder}, law~\cite{law_llm} and medicine~\cite{med_llm}, the interplay among loss prediction and its scaling behavior with model size, and the number of training tokens is yet to be fully explored.
Additionally, the composition of continual pre-training data is simply set up in a heuristic manner~\cite{law_llm,med_llm}, far from being principled. An inappropriate mixture ratio can lead to inefficient training (requiring excessive computation to adapt to specific domains) or insufficient training (failing to adequately reduce domain-specific loss). In light of this, three question hurdles we need to cross are as follows:

{\it Does the optimal data mixture ratio exist for CPT? If so, how does it evolve with model scale or training token volume? Are there any involved simple yet principled laws?}

Currently, several studies examine the scaling laws associated with different data mixture ratios. For instance, \citet{pujianglab_mixing_law} investigate how data mixtures shape scaling laws in the pre-training phase from the ground up, while \citet{ali_cpt_scaling_law} seek to pinpoint the optimal data mixture ratio in CPT, but overlook its crucial connection with the essential trade-off between general and domain loss in CPT.

Therefore, to strengthen our understanding about CPT and guide the experiments in the future, we attempt to address these questions with empirical studies on CPT of LLMs. Specifically, we pre-train several LLMs with different model sizes from scratch and perform CPT on downstream domains (\textit{Finance} and \textit{Academic Papers}) with different data-mixture ratios. Our main contributions can be summarized as follows:

\textbf{Formalization of the Trade-Off in CPT}. We formalize the balance between domain-specific and general abilities during CPT by introducing the concept of feasible mixture ratios. CPT under feasible mixture ratios maintains performance on general data while enhancing performance on domain-specific data. We identify the maximum feasible mixture ratio as the \textbf{Critical Mixture Ratio~(CMR)}, and regard it as the optimal mixture ratio by our definition.

\textbf{Predictability of CMR}. Through extensive experiments, we identify a power-law relationship between loss and both data-mixture ratio and training tokens. As such, we propose CMR scaling law to predict the best mixture ratio by scaling training token volume, which appears to be generalizable based on our findings.

\textbf{Significance of CMR Scaling Law}.
CMR scaling law for CPT is crucial for efficient domain transfer for LLMs. This law allows us to determine the most efficient training configuration by predicting CMR using limited data and compute resources. The finding provides insights into the dynamics of CPT and may offer practical guidelines for optimizing LLM training in specialized domains.

\begin{figure*}[t]
  \includegraphics[width=2\columnwidth]{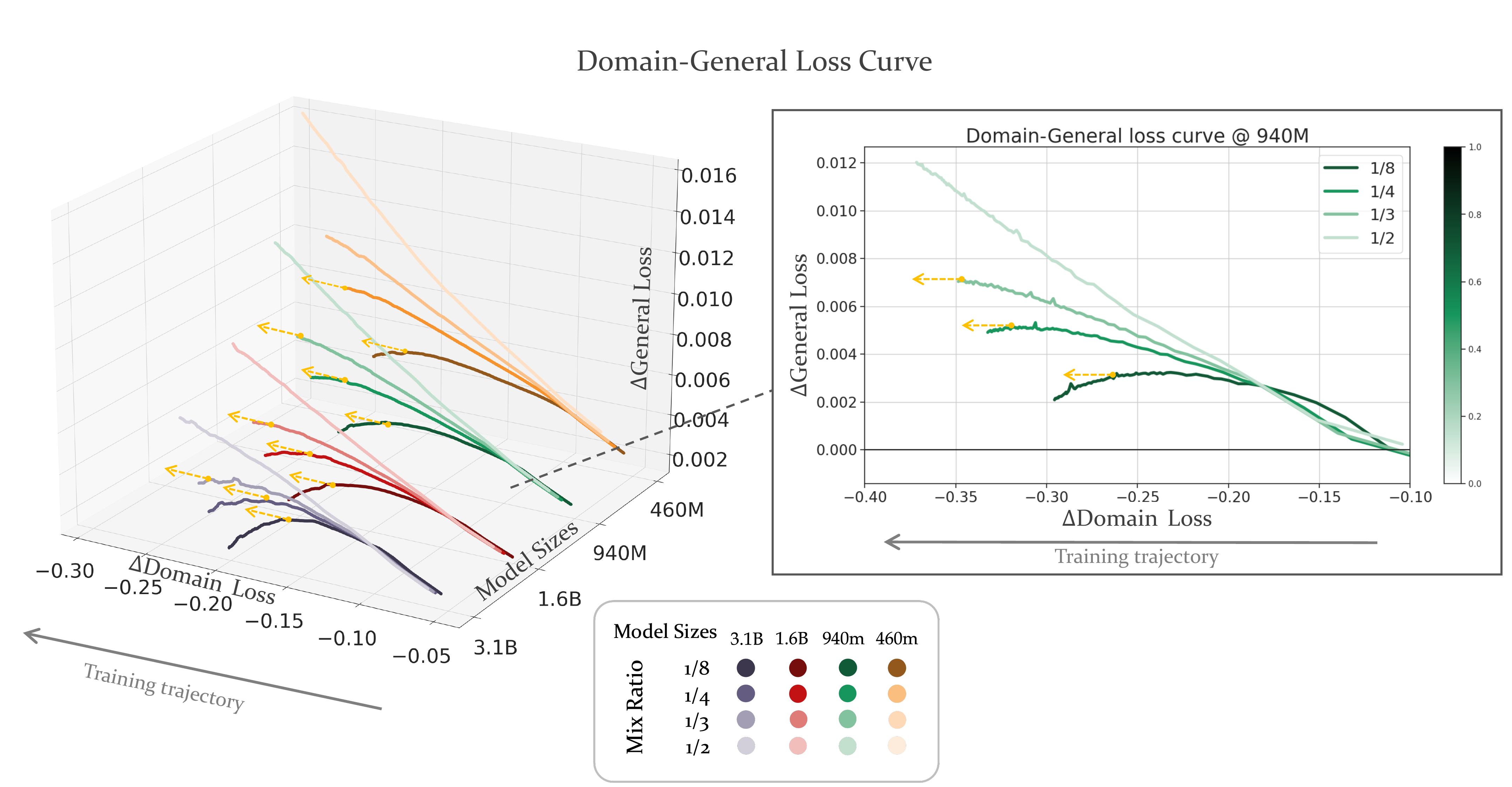}
  \caption{Follow the direction of the training trajectory to track the trend of the curve. Each bunch of lines represents a model size scale: $\{3.1\mathrm{B},1.6\mathrm{B},940\mathrm{M},460\mathrm{M} \}$ and each group of line colors represents the mixture ratios $\{1/8, 1/4, 1/3, 1/2\}$ from dark to light. In order to better display the trend, we have omitted proportions greater than $1/2$. The yellow dashed lines \protect\tikz[baseline=-0.5ex]{
  \protect\draw[thick,dashed, -{Triangle[open, width=6pt, length=4pt]}, line width=0.3mm] 
    (0.5,0) -- (0,0)} point horizontally, indicating the corresponding ratios where $d \mathcal{L}_{\Delta \text{gen}}/d\mathcal{L}_{\Delta \text{dom}}$ closed to $0$.  The third set of lines of model size $940\textrm{M}$, which has been zoomed in and depicted on the right side, showing the trend of the training curve more apparently. All horizontal and vertical cross-sections of the 3D diagram on the left side are detailed in the Appendix~\ref{apd:more_figure}.}
  \label{fig:3d-model}
\end{figure*}

\section{Key Results} 

We train a series of LLMs with multiple mixture ratios of domain-specific data and general data to analyse the scaling behaviour in CPT. The method is detailed in \S~\ref{sec:method}. Based on our experimental setup, we summarize the key results as follows:

\begin{enumerate}
\item The trade-off between two goals of CPT (Definition~\ref{def:CPT object}) suggests that, given a model of certain size, there exists a set of feasible mixture ratios~(Definition~\ref{def:FMR})  that achieve the goals under specific training data constraints. 

\item Basically, general losses in CPT increase initially before decreasing, whereas domain losses tend to decrease. The relationships between loss and mixture ratio, as well as training volume, fit well with a power-law form, allowing for loss prediction under different mixture ratios and training tokens.

\item Using the loss prediction by mixture ratio and training volume, we can predict the CMR (Definition~\ref{def:CMR}) with CMR scaling law. \begin{itemize} 
\item Given the maximum amount of training tokens, experiments in Figure~\ref{fig:3d-model} and predicted results in Figure~\ref{fig:T_0_Predict_R} both show that CMR goes up with increasing model scale: from $29.8\%$ for the 460M model to $34.9\%$ for the 940M model. 
\item CMR depends on the similarity between the target domain and the general domain. The smaller the distribution gap between the two, the larger the CMR. 
Because \textit{Academic Papers} constitute a larger portion of the general data than \textit{Finance},  the pre-trained 460M model tends to show a higher CMR on \textit{Academic Papers} ($36.7\%$) compared to \textit{Finance} ($29.8\%$) during CPT, as illustrated in Figure~\ref{fig:T_0_Predict_R} and Figure~\ref{fig:arxiv_T0_R}.
\end{itemize}
\end{enumerate}
\section{Background and Methods}
The scaling law in the pre-training stage has been widely studied. In this work, we simplify the form of scaling law as much as possible,  which is essentially consistent with previous works in \S~\ref{sec:related work}.

In this section, we will elaborate on the three main concepts involved in this work, including objective of CPT (Definition~\ref{def:CPT object}), feasible mixture ratio (Definition~\ref{def:FMR}) and CMR (Definition~\ref{def:CMR}). Then, we describe our experiment setups, including data preparation, experiment procedures and evaluation.

\subsection{Continual Pre-training on Mixed Dataset} 

\begin{definition}\label{def:CPT object}
Objective of CPT
\end{definition}
Given the pre-trained LLM $M_S$ of model size $S$, general dataset \( \mathcal{D}_{\text{gen}} \), and domain-specific dataset \( \mathcal{D}_{\text{dom}} \), we  continually pre-train $M_S$ on a mixed dataset $D_R$, where the \textbf{mixture ratio} of the domain-specific data is \( R \), with $ R \in [0,1]$. The mixed dataset \( \mathcal{D}_{{R}} \) is denoted as $\mathcal{D}_{R} = \mathcal{D}_{\text{dom}} + \mathcal{D}_{\text{gen}}$ and $R = \frac{|\mathcal{D}_{\text{dom}}|}{{(|\mathcal{D}_{\text{gen}}|+|\mathcal{D}_{\text{dom}}|)}}$.

We define $\mathcal{L}_{\text{gen/dom}}(M_S)$ as the domain or general loss of the model $M_S$. 
We denote $\mathcal{L}^{\text{CPT}}_{\text{gen/dom}}(M_S,\mathcal{D}_{{R}},T)$ as the domain/general loss of model $M_S$ after CPT on dataset $\mathcal{D}_R$ with training token volume $T$. Note that all losses mentioned above are validation losses. The goals for CPT are formalized as follows:


1. By the end of training, the general loss is supposed to either reach plateau or head downward (within a certain tolerance $\epsilon >= 0$):
\begin{equation}
\mathcal{L}^{\text{CPT}}_{\text{gen}}(M_{S},\mathcal{D}_{R},T_{\text{max}}) \leq \mathcal{L}_{\text{gen}}(M_{S}) + \epsilon.
\label{eq:target1}
\end{equation}

2. Domain-specific loss should decline largely:
\begin{equation}
\mathcal{L}^{\text{CPT}}_{\text{dom}}(M_{S}, \mathcal{D}_{R},T_{\text{max}}) < \mathcal{L}_{\text{dom}}(M_{S}).
\label{eq:target2}
\end{equation}

The increase in \( T \) from $0$ to $T_{\text{max}}$ corresponds to the progression of the training trajectory. To better integrate these two aspects, we adopt the method of Lagrange multipliers \cite{lagrange}. 
The loss function \( {F}(\cdot) \) for the whole objective of CPT is the Lagrangian as follows:

\begin{equation}
\begin{aligned}
F(S, R, T, \lambda)& = \mathcal{L}^{\text{CPT}}_{\text{dom}}(M_{S}, \mathcal{D}_{R}, T) \\ 
&+ \lambda ( \mathcal{L}^{\text{CPT}}_{\text{gen}}(M_{S}, \mathcal{D}_{R}, T)\\
&- \mathcal{L}_{\text{gen}}(M_{S}) - \epsilon),
\end{aligned}
\label{eq:opt_function}
\end{equation}
where \( \lambda \) is the Lagrange multiplier used to enforce the constraint on the general loss while minimizing the domain-specific loss. In practice, $\lambda$ governs the importance of two target dimensions in CPT. $F(S, R, T, \lambda)$ is the whole objective function.

Under resource constraints, the optimal training configuration should minimize \( \mathcal{L}_{\text{dom}} \) while satisfying the constraint on \( \mathcal{L}_{\text{gen}} \), which involves finding the optimal \( S \), \( R \), and \( T \) by solving the following optimization problem:
\begin{equation}
\begin{aligned}
&S^*, R^*, T^* = \text{argmin}_{M_S, R, T} \, F(S, R, T, \lambda), \\
\text{s.t.} \ \ 
&\begin{cases} 
\mathcal{L}^{\text{CPT}}_{\text{gen}}(M_S, R, T_{\text{max}}) \leq \mathcal{L}_{\text{gen}}(M_S) + \epsilon, \\
R \geq 0, T \geq 0, \lambda \geq 0.
\end{cases}
\end{aligned}
\label{optimal_with_st}
\end{equation}

\begin{definition}\label{def:FMR}
Feasible Mixture Ratio
\end{definition}
Given fixed model size \( S \), the optimization problem in Equation~(\ref{eq:opt_function}) can be boiled down to $ F(R, T, \lambda)$. We first introduce a mixture ratio set $\mathbb{A}$: according to the first constraint of Definition~\ref{def:CPT object}, under a certain tolerance $\epsilon$ for the deterioration in the final general performance, we can choose a set of mixture ratios $\mathbb{A}$ satisfying $\mathbb{A} = \{ R \mid \mathcal{L}^{\text{CPT}}_{\text{gen}}(M_S, R, T_{\text{max}}) \leq \mathcal{L}_{\text{gen}}(M_S) + \epsilon \}$. Ratios in $\mathbb{A}$ that align with our CPT objective are considered as feasible mixture ratios, denoted as the set \( \mathbb{F} \). A detailed definition transformation is presented in Appendix~\ref{apd:math}, and here we directly provide the formula and the results of derivation: 
within the feasible mixture ratios, there exists a point $T_0$ over the training trajectory of CPT. As CPT proceeds with $T > T_0$, we have $\mathbb{F} = \{R \mid \exists \, T_0 \in (0, T_{\text{max}}) \, : \, \frac{\partial F}{\partial T} \leq 0, R \in \mathbb{A} \}.$

An equivalent condition of defining \(\mathbb{F}\) can be derived as:
\begin{equation}
\begin{aligned}
\mathbb{F} = &\{R \mid \exists \, T_0 \in [0, T_{\text{max}}] \, \\
&: \,\left. \frac{\partial \mathcal{L}_{\Delta\text{gen}}(R, T)}{\partial \mathcal{L}_{\Delta\text{dom}}(R, T)} \right|_{R} = -\frac{1}{\lambda} < 0, R \in \mathbb{A} \}.
\end{aligned}
\label{eq:lambda_derivation}
\end{equation}
For simplicity, we have defined $\mathcal{L}_{\Delta\text{dom}}=\mathcal{L}^{\text{CPT}}_{\text{dom}}-\mathcal{L}_{\text{dom}}$ and $\mathcal{L}_{\Delta\text{gen}}=\mathcal{L}^{\text{CPT}}_{\text{gen}}-\mathcal{L}_{\text{gen}}$.

\paragraph{Visualization} As shown in Figure~\ref{fig:3d-model}, the training curves meeting the objective of CPT are marked with yellow dotted arrows, indicating the curves show a downward trend as training proceeds. The domain loss continuously decreases~({$\mathcal{L}_{\Delta \text{dom}} \downarrow $}) and the general loss is bounded~($d \mathcal{L}_{\Delta \text{gen}}/d\mathcal{L}_{\Delta \text{dom}} \rightarrow 0$) along the training trajectory until the ends of training. This visual representation effectively illustrates the behavior described by Equation~(\ref{eq:lambda_derivation}), demonstrating the trade-off relationship between the domain loss and the general loss during training. The specific derivation and the interpretation of the slope for Figure~\ref{fig:3d-model} is detailed in Appendix~\ref{apd:math}.

\begin{definition}\label{def:CMR}
Critical Mixture Ratio (CMR)
\end{definition}
Given limited compute resources and fixed model size, we hope that the language model can digest domain knowledge more efficiently by achieving the objective as described in Definition~\ref{def:CPT object}. Therefore, we define the maximum among feasible mixture ratios as the Critical Mixture Ratio (CMR)  $R^* = \max \{ R|R \in \mathbb{F}\}$.

The rationale is straightforward: if the ratio is less than CMR, the domain data is not sufficiently utilized in CPT; otherwise, the expected objective can't be achieved, which is manifested as a intolerable increase in general loss, leading to degradation in general ability. Thus, we argue that the CMR is the most suitable ratio for CPT due to the ideal balance of two sides.

\subsection{Method}~\label{sec:method}
\textbf{Data preparation}\quad Our general pre-training data is composed of corpora from Chinese, English, and code. The Chinese corpus and English corpus both include articles from encyclopedia, books, news, papers and social media sites. The code corpus is a subset sampled from StarCoder~\cite{starcoder}.
The general pre-training dataset comprises a total of 220 billion tokens. The proportions of Chinese, English, and code are roughly $44\%:36\%:20\%$.

We meticulously craft two specific domain datasets for CPT: \textit{Finance} and \textit{Academic Papers}. The Finance dataset include financial news, financial policies and regulations, company announcements and research reports from securities and fund companies. The Academic Papers exclusively include papers from Arxiv. 
Each of the datasets contains at least 20 billion tokens, which is sufficient for our CPT.

Unless stated explicitly, all the following results are based on experiments with \textit{Finance}. The results of CPT on \textit{Academic Papers} are reported in \S~\ref{subsec:predicting_cmr}.

\paragraph{LLM Architecture} The involved LLMs in this study have the same architecture as Llama series~\cite{llama,llama2} with standard multi-head attention. The number of parameters ranges from 460M to 3.1B. The architecture is detailed in Table \ref{configuration_llms} of Appendix.
\paragraph{Experiment Setup}
We split the general pre-training dataset into two subsets: a 200B-token general dataset for general pre-training and a 20B-token general dataset for CPT.

In the pre-training stage, we pre-train the LLMs from scratch with 200B-token general dataset with a max learning rate of 3e-4, a batch size of 512, and a sequence length of 4096. The training step is 100,000 for each LLM. In the CPT stage, we train each LLM for another 10,000 steps (20 billion tokens) with a max learning rate of 3e-5 and warmup-constant LR schedule, on a mixture of the 20B-token general dataset and a domain dataset with different mixture ratios.

\paragraph{Evaluation}

Scaling laws emphasize the predictability of pre-training loss~\cite{openai_scaling_law,chinchilla_scaling_law,rl_openai_scaling_law,openai_sft_scaling_law}, which is a widely-used performance indicator. Recent studies~\cite{understanding_loss,scaling_relationship_loss_ali} highlight that pre-training loss is highly correlated with downstream task performance. Therefore, we use the pre-training loss on the validation set to measure the model's capability of general or domain-specific task during the CPT process. In addition, we use Mean Squared Error (MSE) and R-square ($R^2$) to measure the quality of the fitting, which provides a clear and interpretable analysis of the errors.

\section{Does the Critical Mixture Ratio Exist?}
\begin{flushright}
\textemdash\textemdash\ Yes, the CMR does exist.
\end{flushright}
A larger mixture ratio implies a higher proportion of domain-specific data in the training set, resulting in a lower domain loss. However, due to the potential catastrophic forgetting of domain transfer, it is essential to ensure that the loss in the new domain continues to decrease while the original capabilities of LLMs are preserved and not compromised during CPT. 
Consequently, a higher mixture ratio is not always best. This raises an important question: does a Critical Mixture Ratio (CMR) exist that can balance these two goals of CPT in Definition~\ref{def:CPT object} effectively and efficiently?

Figure~\ref{fig:3d-model} (left) demonstrates that for models of various sizes, there is at least one curve at a specific ratio that shows a downward trend, highlighted by yellow dotted arrows. This indicates the presence of feasible mixture ratios that align with our CPT objective. On the other hand, larger models tend to have bigger feasible mixture ratios set $\mathbb{F}$~(more curves with yellow dotted arrows). For curves that meet the objective of CPT, a higher ratio is preferable, as it incorporates more domain knowledge while optimizing training efficiency within the tolerance of decline in general capacity. Therefore, the critical mixture ratio is defined as the highest proportion among these satisfactory curves, representing the optimal ratio for the given model size and limited training token volume. 

If feasible ratios exist, we can conclude that CMR is also supposed to exist. Fundamentally, the existence of CMR arises from the trade-off between general and domain-specific capabilities, as well as the limited data and computing resources. According to definition ~\ref{def:CMR}, the CMR is present across models of different scales, as shown in Figure~\ref{fig:3d-model}. This figure illustrates the existence of CMR as the maximum value within the feasible set. However, the precise value of CMR can not be determined from the figure, as it requires extensive experiments with different mixture ratios. The estimation of CMRs is discussed in ~\ref{subsec:predicting_cmr} and plotted in Figure~\ref{fig:T_0_Predict_R}.

To look closely, we enlarged the longitudinal section of $M_{940\textrm{M}}$ in the 3D graph in Figure~\ref{fig:3d-model} and placed it on the right side. It can be seen that as the mixture ratio increases, the curve continues to rise until the loss in the general domain exceeds our tolerance. The potentially controversial issue is that the downward trend in one-third of the curves is as clear as in the rest. The reason why it is feasible curve here can be found in Appendix~\ref{apd:math}. Although it is not easily noticeable, there are indeed points on this curve where the slope is less than $0$.

\begin{figure}[h]
  \includegraphics[width=1.0\columnwidth]{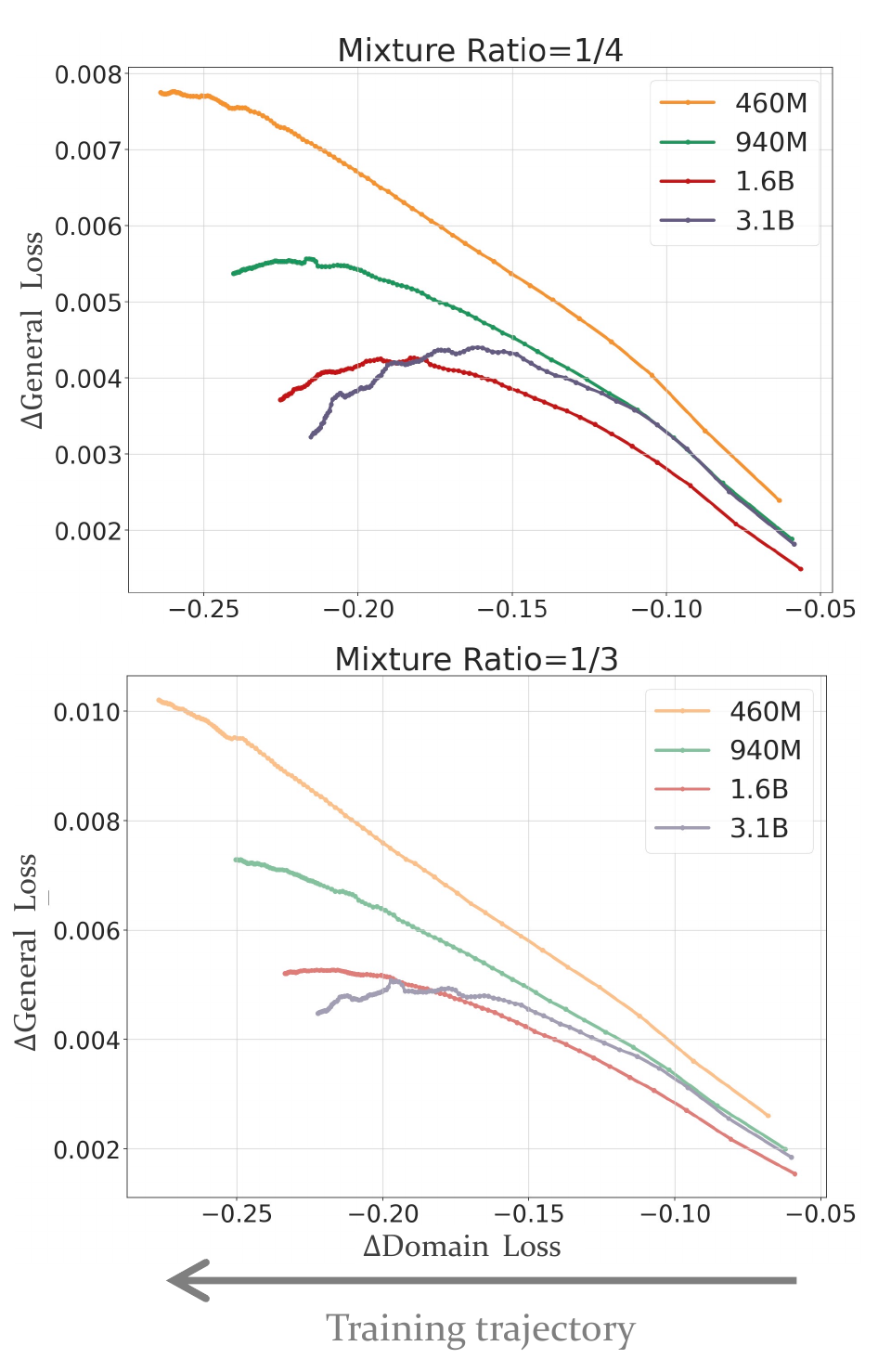}
  \caption{Follow the direction of the training trajectory to track the trend of the curve. The $\mathcal{L}_{\Delta \text{gen}}$ and $\mathcal{L}_{\Delta \text{dom}}$ loss functions for the models at mixture ratios of $1/4$ and $1/3$ are illustrated.}
  \label{fig:two_ratio}
\end{figure}

\paragraph{Findings}
From another perspective, we plot the loss curves of the models under the same mixture ratio as shown in Figure~\ref{fig:two_ratio}. When the mixture ratio is $1/4$, all models can achieve the training objective of CPT. However, at a $1/3$ mixture ratio, only $M_{940\textrm{M}}$, $M_{1.6\textrm{B}}$ and $M_{3.1\textrm{B}}$ achieve the CPT goal. This indicates that CMR for $M_{940\textrm{M}}$ is around $1/4$ within the scope of our training token volumes, while the CMRs for $M_{940\textrm{M}}$, $M_{1.6\textrm{B}}$ and $M_{3.1\textrm{B}}$ are at least $1/3$. In other words, \textbf{CMRs slightly increase with model size, suggesting that larger models can accommodate a higher proportion of domain data.} We also further this finding by taking more cross-sections of Figure~\ref{fig:3d-model}~(left) in Appendix~\ref{apd:more_figure} and the predicted CMR in following \S~\ref{sec:RQ2}.

This phenomenon can be explained by the models' ability to consume domain knowledge. As the proportion of domain-specific data increases, the knowledge that the model needs to learn also increases. LLMs with smaller size struggle to absorb much of domain knowledge while preserving the general knowledge, leading to a degradation in their original general performance. In contrast, models with larger sizes can accommodate more knowledge with more parameters, thereby maintaining better performance.

\section{Is CMR Predictable? }\label{sec:RQ2}
\begin{flushright}
\textemdash\textemdash\ Yes, the CMR can be predicted.
\end{flushright}
The existence of CMR indicates that in the process of CPT, we may explore the CMR scaling law to seek the best mixture ratio under resource constraints and domain data limitations, thereby optimizing training effectiveness and efficiency. In other words, the next question to answer is whether we can predict the CMR for model $M_s$ given a maximum amount of continuation training token volume, $T_{\textrm{max}}$.

To this end, two basic prerequisites must be met: predicting losses for different mixture ratio and predicting losses for different training token volume. In this section, we will demonstrate that these two prerequisites have been satisfied separately in \S~\ref{subsec:loss_and_r} and \S~\ref{subsec:loss_and_token}, and finally detail the scaling law to predict CMR in \S~\ref{subsec:predicting_cmr}. To keep notations simple, we omit fixed variables in the loss function ($\mathcal{L}_\text{dom/gen}$ and $\mathcal{L}_{\Delta \text{dom}/\Delta \text{gen}}$) in this following.

\begin{figure}[t]
  \includegraphics[width=0.95\columnwidth]{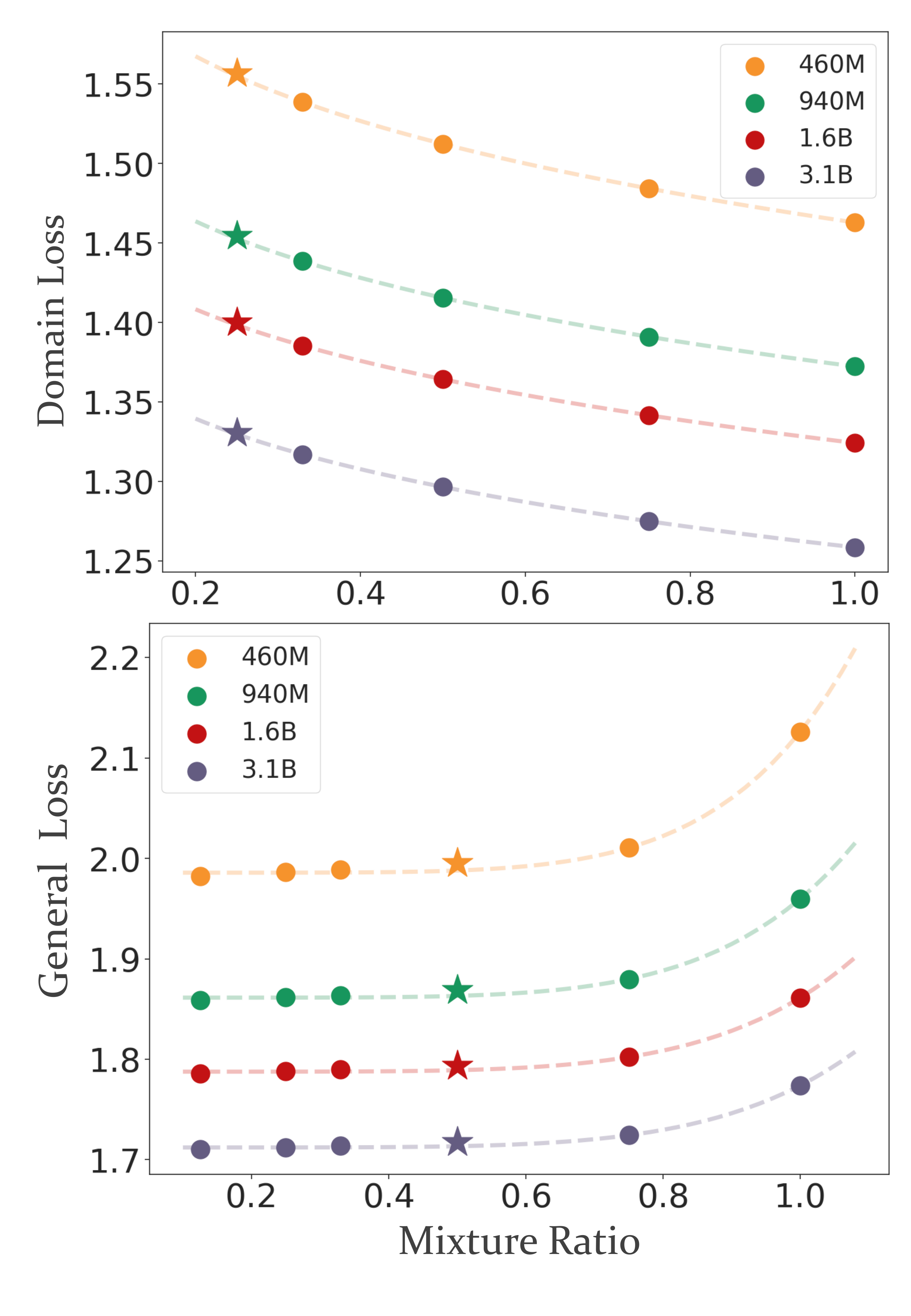}
  \caption{The upper figure shows the fitting curve of domain loss $\mathcal{L}_\text{dom}$ with the change of mixture ratio $R$, and the lower figure shows the fitting curve of general loss $\mathcal{L}_\text{gen}$. The solid circles~($\bullet$) represent real losses, and the stars~(\ding{72}) represent the predicted losses.}
  \label{fig:gen-dom-ratio-predict}
\end{figure}

\subsection{Predicting Losses of Mixture Ratio}\label{subsec:loss_and_r}

Predicting the general and domain loss is closely related to understanding the scaling behavior in the CPT stage. We study the scaling behavior of losses at $T=T_{\mathrm{max}}$. In addition, since scaling law aims to fit data points, their parametric forms should be intrinsically related to the observed trends in the data points. Based on previous works~\cite{openai_scaling_law,chinchilla_scaling_law} and data trends we observed, we proposed the simplified expression
\({\mathcal{L}(R)}\) as a power-law form of 
$$
\mathcal{L}(R) = \alpha \cdot R^{s} + \beta,
$$
where $\alpha$ is a coefficient, $s$ is the exponent, and $\beta$ is the bias. 

As shown in Figure~\ref{fig:gen-dom-ratio-predict}, domain loss gradually decreases with the increase of the mixture ratio, while general loss remains almost unchanged initially and then begins to rise. After fitting the general loss and domain loss separately for different mixture ratios $R$ (non-endpoint values, $R \in (0,1)$), we make predictions on new ratios. As shown in Figure~\ref{fig:gen-dom-ratio-predict}, the predicted values align closely with the fitted curve. Notably, the predictions demonstrate high accuracy, with error values within $0.05\%$ as presented in Table~\ref{tab:domain_ratio_predict}.

Given the predicted $\mathcal{L}_{\text{dom}}(R)$ and $\mathcal{L}_{\text{gen}}(R)$ under different mixture ratios, we can obtain a range of mixture ratios that fulfil the tolerance limit $\epsilon$, denoted as~$\mathbb{A}$, according to Equation~\ref{eq:opt_function}. In the objective of CPT we set, $\epsilon = 0.05$.

\subsection{Predicting Losses of Training Tokens}\label{subsec:loss_and_token}

\begin{figure}[t]
  \includegraphics[width=\columnwidth]{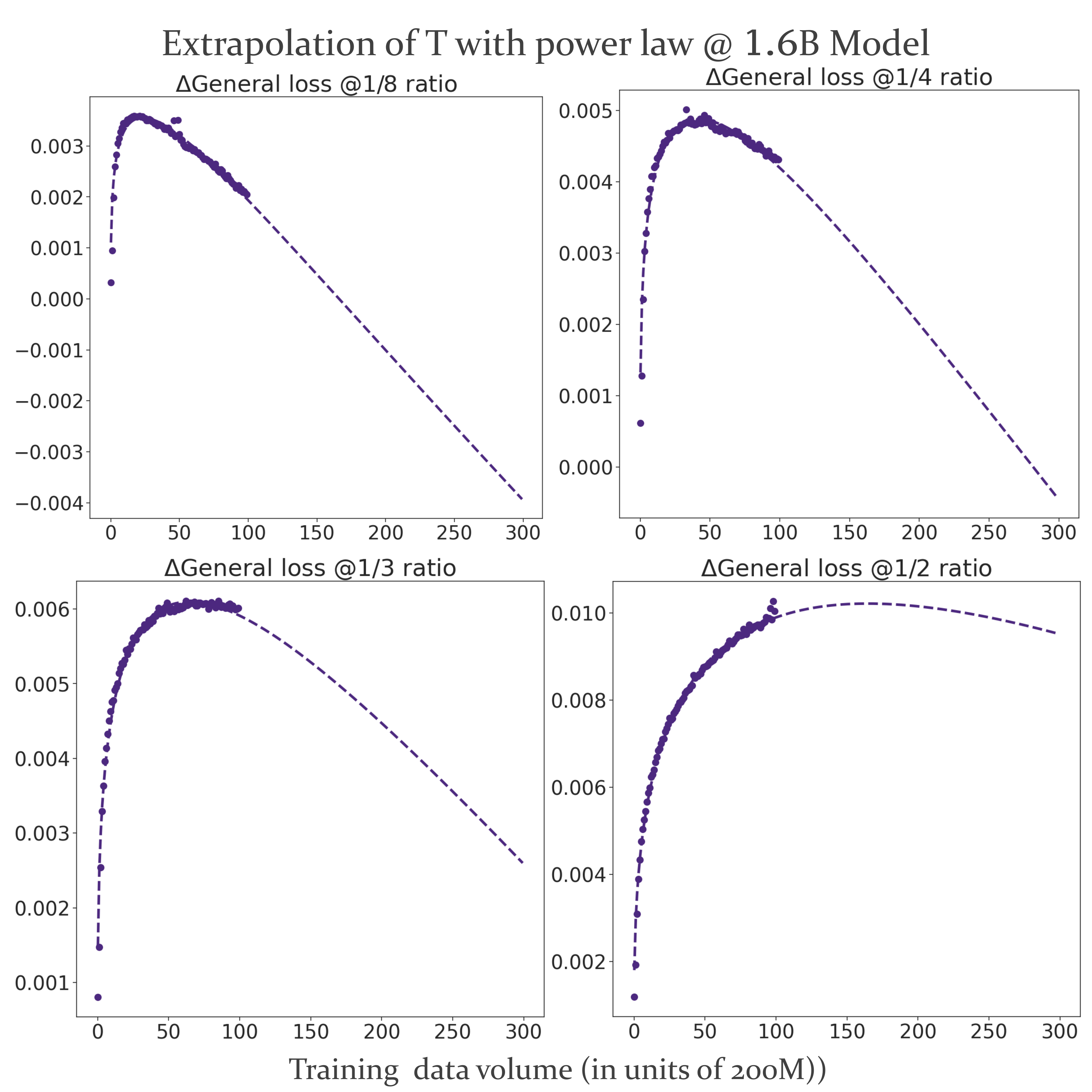}
  \caption{The figure shows the general loss of $M_{1.6B}$ fitting and extrapolating at four distinct mixture ratios: $\{1/8, 1/4, 1/3, 1/2\}$. As the ratio increases, the curve gradually rises when training data volume increases.}
  \label{fig:predicting_T}
\end{figure}

Previous works~\cite{openai_scaling_law,chinchilla_scaling_law} have shown that the model size $S$ and the volume of training tokens $T$ can be used to fit the power law of loss. However, our work differs in two key aspects. First, we model the change of loss $\mathcal{L}_{\Delta \text{dom}/\Delta \text{gen}}(T)$ rather than the loss itself. Second, due to the phenomenon of general loss initially increasing and then decreasing as shown in Figure~\ref{fig:predicting_T}, we leverage a two-term polynomial function for better fitting. According to Equation~\ref{eq:opt_function_delta} in Appendix~\ref{apd:math}, the loss for CPT training tokens $T$ is formulated as follows: 

\begin{equation}
\left\{
\begin{aligned}
\mathcal{L}_{\Delta \text{dom}}(T) &= \alpha_1 \cdot T^{s_1} + \beta_1 ,\\
\mathcal{L}_{\Delta \text{gen}}(T) &= \alpha_2 \cdot T^{s_2} + \alpha_3 \cdot T^{s_3} + \beta_2.
\end{aligned}
\right.
\label{eq:dom-gen-token-law}
\end{equation}
where \(\alpha_1\), \(\alpha_2\), \(\alpha_3\), \(\beta_1\), \(\beta_2\), \(s_1\), \(s_2\), and \(s_3\) are learnable parameters. Our results demonstrate that the form \eqref{eq:dom-gen-token-law} exhibits high fitting accuracy with low MSE and high $R^2$ in Table~\ref{tab:MSE_R2} and Figure~\ref{fig:predicting_T}.

\subsection{Predicting CMR}\label{subsec:predicting_cmr}

\begin{figure}[t]
  \includegraphics[width=\columnwidth]{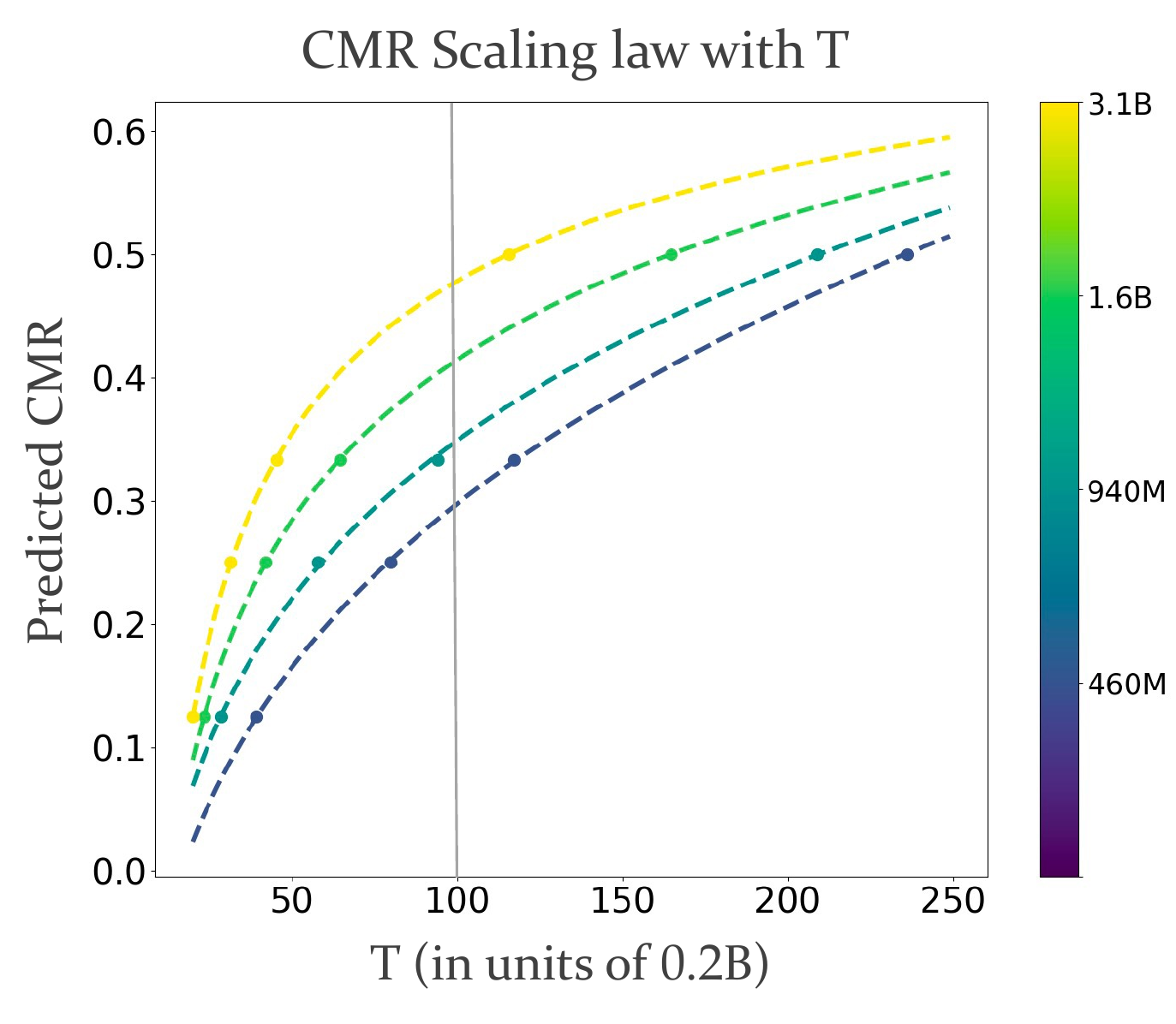}
  \caption{We can use the CMR scaling laws to predict CMRs under fixed model size $S$, and are extrapolated to $T=250$, which is equivalent to a training volume of $500\mathrm{B}$ tokens.}
  \label{fig:T_0_Predict_R}
\end{figure} 

According to the definition of feasible mixture ratios in Definition~\ref{def:FMR} and the method for determining the set $\mathbb{F}$ in Appendix~\ref{apd:math}, where $\mathbb{F} \subset \mathbb{A}$, and $\mathbb{A}$ is obtained by predicting losses for any mixture ratio in \S~\ref{subsec:loss_and_r}, we can establish a relationship between training token volume $T$ and the feasible mixture ratios by the fitting laws in \S~\ref{subsec:loss_and_token}. Overall, based on the parameters provided in Formula~\ref{eq:dom-gen-token-law}, the critical solution $T_0$ is obtained for a specific mixture ratio $R_0$ denoted as (derivation detailed in Appendix~\ref{apd:math}):
\begin{equation}
\begin{aligned}
&T_0 =\\
& \left[-\frac{\alpha_1 \cdot s_1}{\lambda \alpha_2 \cdot s_2} \left(1 + \frac{\alpha_3 \cdot s_3}{\alpha_2 \cdot s_2} T_0^{s_3 - s_2}\right)^{-1}\right]^{\frac{1}{s_2 - s_1}}
\end{aligned}
\end{equation}
When $T_0$ is less than the given maximum training token volume $T_{\mathrm{max}}$, we can conclude that the current ratio $R_0$ is a feasible mixture ratio. Conversely, if $T_0$ exceeds $T_{\mathrm{max}}$, then $R_0$ is not a feasible mixture ratio. If $T_0$ is equal to $T_{\mathrm{max}}$, then $R_0$ is the critical ratio. We propose the following CMR scaling law: 
\begin{equation}
R_{\textrm{CMR}} = \alpha_4 \cdot T^{s_4} + \beta_3.
\end{equation}
The fitting curves are showed in Figure~\ref{fig:T_0_Predict_R}. 
In our experiments, $T_{\mathrm{max}}$ is $20\mathrm{B}$ tokens, which corresponds to a value of $T=100$ in the figure. Therefore, for four models of different scales, their predicted CMR are $ 29.8\%, 34.9\%, 41.4\%$ and $47.8\%$ for $M_{460\mathrm{M}},M_{940\mathrm{M}},M_{1.6\mathrm{B}},M_{3.1\mathrm{B}}$, respectively.

\paragraph{Generalization} 

\begin{figure}
  \raggedright
  \includegraphics[width=0.9\columnwidth]{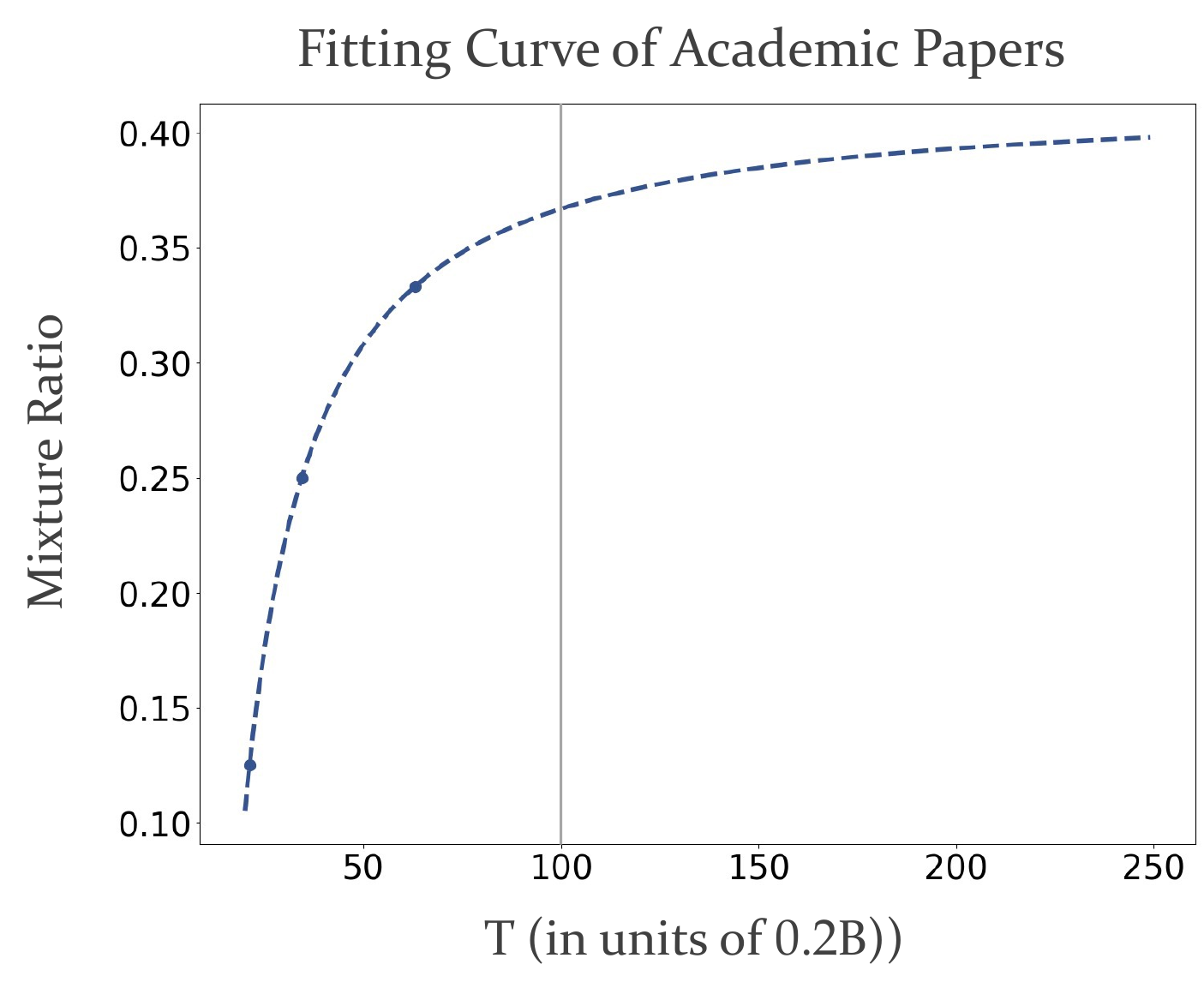}
  \caption{With a fixed model size $S=460\textrm{M}$, using the CMR scaling law can be extrapolated to $T=250$ and more. We can use the CMR scaling laws to predict CMR for \textit{Academic Papers} in the CPT of $M_{460\textrm{M}}$. When $T=T_{\mathrm{max}}=100$, the value of $R$ is $36.7\%$, regarded as the CMR.}
  \label{fig:arxiv_T0_R}
\end{figure}

\begin{figure*}[t]
  \includegraphics[width=2\columnwidth]{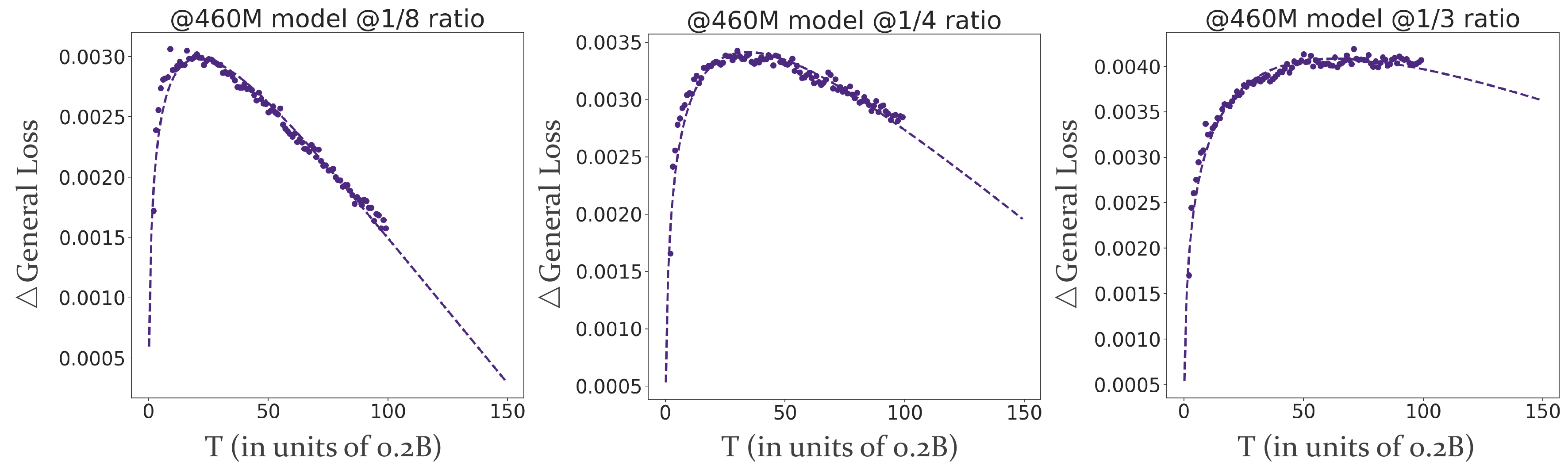}
  \caption{The figure shows the general loss of $M_{460\textrm{M}}$ fitting and extrapolating at three distinct mixture ratios: $\{1/8, 1/4, 1/3\}$ with CPT on \textit{Academic Papers}.}
  \label{fig:arxiv_T_prediction}
\end{figure*}

In order to verify whether the CMR scaling law can be generalized, we experiment on another domain \textit{Academic Papers} with different mixture ratios. In this generalization experiment, we only conduct CPT on the 460M-sized model with {Academic Papers} data proportions set to $\{1/8, 1/4, 1/2, 3/4, 1/3\}$ respectively. All other settings were kept consistent with \textit{Finance}. As shown in  Figure~\ref{fig:arxiv_T_prediction}, the trade-off of CPT still exist in this domain, and thus there exists a CMR. Furthermore, the CMR scaling law still work, which can observed in Figure~\ref{fig:arxiv_T0_R}. The predicted CMR for \textit{Academic Papers} is $36.7\%$, given the maximum training token volume $T_{\textrm{max}} = 100$.

\paragraph{Finding}
Comparing CMR predictions in two different domains, we find that the pre-trained model of the same size (460M) shows a higher CMR for \textit{Academic Papers} ($36.7\%$) compared to \textit{Finance} ($29.8\%$) during CPT. As illustrated in \S~\ref{sec:method}, we performed a statistical analysis on the general pre-training dataset, finding that data from the \textit{Academic Papers} domain accounts for about $10\%$, while the \textit{Finance} corpus is less than that. In other words, \textbf{the smaller the distribution gap between the target and general domains, the larger the CMR.} This observation aligns with related research findings~\cite{ke2022continual,ke2023cpt} and can be attributed to the difficulty of domain adaptation. When the distribution gap between the target and general domains is smaller, domain adaptation becomes easier, reducing the risk of degradation of general performance during CPT, even with a higher proportion of domain-specific data.




\paragraph{Open Discussion}

As showed in Figs. ~\ref{fig:T_0_Predict_R} and ~\ref{fig:arxiv_T0_R}, the larger the $T_{\text{max}}$, the wider the range of feasible mixture ratios. Therefore, it seems that when $T_{\text{max}}$ tends to be infinity (the amount of data available for continued training is infinite, and computational resources are unlimited completely), the range of feasible mixture ratios would approach $(0,1)$, leading CMR approaching $1$. In this sense, each curve of CPT trajectory will show an expected convergence trend of objective, provided that there is enough $T$ to allow it to develop.

Moreover, we find out that the solution of $T_0$ in Definition~\ref{def:FMR} approaches the inflexion point of $\mathcal{L}_{\Delta\text{gen}}(T)$ in Figure~\ref{fig:predicting_T} and Figure~\ref{fig:arxiv_T_prediction}, when $\lambda \rightarrow +\infty$, which we used for solving equations~\ref{eq:partial_derivative_zero} ranges from $100$ to $7000$. The reason is likely to be that, the change in the general loss is much smaller than the change in domain, and $\lambda$ in the objective function of CPT needs to be very large to amplify such subtle changes within the tolerance of constraints. In addition, during the training process, the decreasing trend of domain loss has always been present, but there are obvious inflection points in the  of general loss curve~(rise first and then fall). That is to say, by only locating the inflection points on the general loss curve and finding this distance to the max training token volume, we can roughly estimate how far away we are from CMR at current ratio. 
\section{Related Work}\label{sec:related work}

\subsection{Continual Pre-training}

Continual Pre-Training (CPT) aims to perpetually pre-train large language models (LLMs), allowing them to adapt to new domains and reducing the high costs associated with training models from scratch for specialized tasks \cite{cpt_llm_survey}. CPT can be employed to tailor LLMs for specific fields, such as code \cite{autocoder,starcoder}, medicine \cite{med_llm}, law \cite{law_llm}, and science. By using an appropriate mixture of data from various domains \cite{cpt_med_tradition_nlp}, CPT not only enhances downstream performance but also mitigates the issue of catastrophic forgetting \cite{zhanghengyuan_balancing}, which is prevalent in all forms of post-training \cite{cpt_in_cv,empirical_cv_catastrophic_forgetting}.

Many recent studies on domain-specific LLMs~\cite{med_llm,law_llm} adopt replay strategies (mixing general and domain-specific data) to concern about general losses during CPT. Other works~\cite{ali_cpt_scaling_law,stability_gap,Data_Mixing_Made_Efficient:} have also noted challenges with maintaining general abilities, which aligns with our findings. For example, \citet{stability_gap} identified a \textit{Stability Gap}, where performance initially drops during CPT and then gradually recovers, leading to inefficient pre-training and potential forgetting of general knowledge. 
In our work, we focus on the trade-off between general and domain-specific performance (losses) during CPT. To facilitate clearer experimental observations, we chose to restrict CPT to a single domain and did not explore the more complex setting of training across multiple domains.

\subsection{Scaling Law}
Numerous studies~\cite{deep_learning_scaling_law_baidu,multimodel_scaling_law,explain_scaling_law,openai_scaling_law,chinchilla_scaling_law,nanolm} demonstrate a power-law relationship between performance and the increase in both the number of parameters and the size of the training data.
These relationships are crucial for large language models (LLMs), being of paramount importance in various stages such as pre-training~\cite{openai_scaling_law,chinchilla_scaling_law,pujianglab_mixing_law}, supervised fine-tuning (SFT)~\cite{openai_sft_scaling_law,rectified_sft_scaling_law}, etc. Recently, researchers describe scaling laws from various different perspectives~\cite{data_scaling_law,pujianglab_mixing_law}. 
The form of the scaling law used in this papers is consistent with \citet{chinchilla_scaling_law}, $L = E + \frac{A}{S^\alpha} + \frac{B}{T^\beta}$, where $\{E, A, B, \alpha, \beta\}$ are fitting parameters. However, we express in an simpler and more appropriate way for our demonstrations.

\subsection{Data Mixture Scaling Law}
Several studies have examined the scaling laws associated with various data mixture ratios. For instance, \citet{pujianglab_mixing_law} investigate how different data mixtures influence scaling laws during the pre-training phase. However, their proposed laws are not applicable to CPT. Another study by \citet{ali_cpt_scaling_law} aims to identify the optimal data mixture ratio using the D-CPT law. Their method focuses solely on minimizing domain loss by fixing model sizes and training token volume, thereby neglecting the trade-off between general loss and domain loss, which is critical in CPT.

\section{Conclusion}
In this work, we investigated the scaling behavior of LLMs under Continual Pre-Training (CPT) to address the limitations of domain-specific performance. We provided a clear definition of Critical Mixture Ratio (CMR) for optimizing the mixture ratio of general and domain-specific data. Our experiments revealed a power-law relationship between loss, mixture ratio, and training data size, allowing us to predict the CMR efficiently. These findings may offer practical guidelines for optimizing LLM training, helping to balance general and domain-specific performance while minimizing resource consumption. Additionally, our study suggests that understanding the CPT process and scaling laws could be valuable for future research aimed at enhancing LLM capabilities in specialized fields.

\section{Limitations}
\paragraph{Computational Constraints} We experimented with model sizes range from 400M to 3.1B. However, the largest model in our experiments is still relatively small among contemporary LLMs. It may lead to inaccuracy in estimation of model size scaling.

\paragraph{Limited Domains} In this work, we conducted continual pre-training only on two specific domains (finance and academic papers) respectively. Although we have draw some useful conclusions from the experimental results, experiments with more domains are expected to provide more refined results and likely to bring some new insights.

\paragraph{CMR scaling law with model size} The CMR scaling in this work can only predict the CMR of a fixed model size. We have not explored how to predict CMR of large models with experiments on small models. An possible method is that first we extrapolate all the losses of small models to large models with model size scaling law, and use CMR scaling law to predict the CMR of the large model. We left it as a future work to predict CMR by leveraging multiple scaling laws with less computational efforts.

\paragraph{Downstream Evaluation}
As stated in \S~\ref{sec:method}, our primary focus was on establishing and validating the CMR scaling law through loss metrics, which is widely used and highly correlated with downstream task performance. However, this study did not directly evaluate performance on downstream tasks. Including downstream task performance could provide a more intuitive understanding of the observed trends.

\bibliography{custom}

\begin{thebibliography}{32}
\providecommand{\natexlab}[1]{#1}

\bibitem[{Bahri et~al.(2021)Bahri, Dyer, Kaplan, Lee, and Sharma}]{explain_scaling_law}
Yasaman Bahri, Ethan Dyer, Jared Kaplan, Jaehoon Lee, and Utkarsh Sharma. 2021.
\newblock Explaining neural scaling laws.
\newblock \emph{arXiv preprint arXiv:2102.06701}.

\bibitem[{Brown et~al.(2020)Brown, Mann, Ryder, Subbiah, Kaplan, Dhariwal, Neelakantan, Shyam, Sastry, Askell, Agarwal, Herbert-Voss, Krueger, Henighan, Child, Ramesh, Ziegler, Wu, Winter, Hesse, Chen, Sigler, Litwin, Gray, Chess, Clark, Berner, McCandlish, Radford, Sutskever, and Amodei}]{gpt3}
Tom~B. Brown, Benjamin Mann, Nick Ryder, Melanie Subbiah, Jared Kaplan, Prafulla Dhariwal, Arvind Neelakantan, Pranav Shyam, Girish Sastry, Amanda Askell, Sandhini Agarwal, Ariel Herbert-Voss, Gretchen Krueger, Tom Henighan, Rewon Child, Aditya Ramesh, Daniel~M. Ziegler, Jeffrey Wu, Clemens Winter, Christopher Hesse, Mark Chen, Eric Sigler, Mateusz Litwin, Scott Gray, Benjamin Chess, Jack Clark, Christopher Berner, Sam McCandlish, Alec Radford, Ilya Sutskever, and Dario Amodei. 2020.
\newblock \href {https://arxiv.org/abs/2005.14165} {Language models are few-shot learners}.
\newblock \emph{Preprint}, arXiv:2005.14165.

\bibitem[{Chen et~al.(2023)Chen, Cano, Romanou, Bonnet, Matoba, Salvi, Pagliardini, Fan, K{\"o}pf, Mohtashami et~al.}]{med_llm}
Zeming Chen, Alejandro~Hern{\'a}ndez Cano, Angelika Romanou, Antoine Bonnet, Kyle Matoba, Francesco Salvi, Matteo Pagliardini, Simin Fan, Andreas K{\"o}pf, Amirkeivan Mohtashami, et~al. 2023.
\newblock Meditron-70b: Scaling medical pretraining for large language models.
\newblock \emph{arXiv preprint arXiv:2311.16079}.

\bibitem[{Colombo et~al.(2024)Colombo, Pires, Boudiaf, Culver, Melo, Corro, Martins, Esposito, Raposo, Morgado et~al.}]{law_llm}
Pierre Colombo, Telmo~Pessoa Pires, Malik Boudiaf, Dominic Culver, Rui Melo, Caio Corro, Andre~FT Martins, Fabrizio Esposito, Vera~L{\'u}cia Raposo, Sofia Morgado, et~al. 2024.
\newblock Saullm-7b: A pioneering large language model for law.
\newblock \emph{arXiv preprint arXiv:2403.03883}.

\bibitem[{Cossu et~al.(2022)Cossu, Tuytelaars, Carta, Passaro, Lomonaco, and Bacciu}]{cpt_in_cv}
Andrea Cossu, Tinne Tuytelaars, Antonio Carta, Lucia Passaro, Vincenzo Lomonaco, and Davide Bacciu. 2022.
\newblock Continual pre-training mitigates forgetting in language and vision.
\newblock \emph{arXiv preprint arXiv:2205.09357}.

\bibitem[{Du et~al.(2024)Du, Zeng, Dong, and Tang}]{understanding_loss}
Zhengxiao Du, Aohan Zeng, Yuxiao Dong, and Jie Tang. 2024.
\newblock Understanding emergent abilities of language models from the loss perspective.
\newblock \emph{arXiv preprint arXiv:2403.15796}.

\bibitem[{Gao et~al.(2023)Gao, Schulman, and Hilton}]{rl_openai_scaling_law}
Leo Gao, John Schulman, and Jacob Hilton. 2023.
\newblock Scaling laws for reward model overoptimization.
\newblock In \emph{International Conference on Machine Learning}, pages 10835--10866. PMLR.

\bibitem[{Ge et~al.(2024)Ge, Ma, Chen, Li, and Ding}]{Data_Mixing_Made_Efficient:}
Ce~Ge, Zhijian Ma, Daoyuan Chen, Yaliang Li, and Bolin Ding. 2024.
\newblock Data mixing made efficient: A bivariate scaling law for language model pretraining.
\newblock \emph{arXiv preprint arXiv:2405.14908}.

\bibitem[{Guo et~al.(2024)Guo, Fu, Zhang, Zhao, and Shen}]{stability_gap}
Yiduo Guo, Jie Fu, Huishuai Zhang, Dongyan Zhao, and Yikang Shen. 2024.
\newblock Efficient continual pre-training by mitigating the stability gap.
\newblock \emph{arXiv preprint arXiv:2406.14833}.

\bibitem[{Gururangan et~al.(2020)Gururangan, Marasovi{\'c}, Swayamdipta, Lo, Beltagy, Downey, and Smith}]{cpt_med_tradition_nlp}
Suchin Gururangan, Ana Marasovi{\'c}, Swabha Swayamdipta, Kyle Lo, Iz~Beltagy, Doug Downey, and Noah~A. Smith. 2020.
\newblock \href {https://doi.org/10.18653/v1/2020.acl-main.740} {Don{'}t stop pretraining: Adapt language models to domains and tasks}.
\newblock In \emph{Proceedings of the 58th Annual Meeting of the Association for Computational Linguistics}, pages 8342--8360, Online. Association for Computational Linguistics.

\bibitem[{Henighan et~al.(2020)Henighan, Kaplan, Katz, Chen, Hesse, Jackson, Jun, Brown, Dhariwal, Gray et~al.}]{multimodel_scaling_law}
Tom Henighan, Jared Kaplan, Mor Katz, Mark Chen, Christopher Hesse, Jacob Jackson, Heewoo Jun, Tom~B Brown, Prafulla Dhariwal, Scott Gray, et~al. 2020.
\newblock Scaling laws for autoregressive generative modeling.
\newblock \emph{arXiv preprint arXiv:2010.14701}.

\bibitem[{Hernandez et~al.(2021)Hernandez, Kaplan, Henighan, and McCandlish}]{openai_sft_scaling_law}
Danny Hernandez, Jared Kaplan, Tom Henighan, and Sam McCandlish. 2021.
\newblock Scaling laws for transfer.
\newblock \emph{arXiv preprint arXiv:2102.01293}.

\bibitem[{Hestness et~al.(2017)Hestness, Narang, Ardalani, Diamos, Jun, Kianinejad, Patwary, Yang, and Zhou}]{deep_learning_scaling_law_baidu}
Joel Hestness, Sharan Narang, Newsha Ardalani, Gregory Diamos, Heewoo Jun, Hassan Kianinejad, Md~Mostofa~Ali Patwary, Yang Yang, and Yanqi Zhou. 2017.
\newblock Deep learning scaling is predictable, empirically.
\newblock \emph{arXiv preprint arXiv:1712.00409}.

\bibitem[{Hoffmann et~al.(2022)Hoffmann, Borgeaud, Mensch, Buchatskaya, Cai, Rutherford, Casas, Hendricks, Welbl, Clark et~al.}]{chinchilla_scaling_law}
Jordan Hoffmann, Sebastian Borgeaud, Arthur Mensch, Elena Buchatskaya, Trevor Cai, Eliza Rutherford, Diego de~Las Casas, Lisa~Anne Hendricks, Johannes Welbl, Aidan Clark, et~al. 2022.
\newblock Training compute-optimal large language models.
\newblock \emph{arXiv preprint arXiv:2203.15556}.

\bibitem[{Kaplan et~al.(2020)Kaplan, McCandlish, Henighan, Brown, Chess, Child, Gray, Radford, Wu, and Amodei}]{openai_scaling_law}
Jared Kaplan, Sam McCandlish, Tom Henighan, Tom~B Brown, Benjamin Chess, Rewon Child, Scott Gray, Alec Radford, Jeffrey Wu, and Dario Amodei. 2020.
\newblock Scaling laws for neural language models.
\newblock \emph{arXiv preprint arXiv:2001.08361}.

\bibitem[{Ke et~al.(2022)Ke, Lin, Shao, Xu, Shu, and Liu}]{ke2022continual}
Zixuan Ke, Haowei Lin, Yijia Shao, Hu~Xu, Lei Shu, and Bing Liu. 2022.
\newblock Continual training of language models for few-shot learning.
\newblock \emph{arXiv preprint arXiv:2210.05549}.

\bibitem[{Ke et~al.(2023)Ke, Shao, Lin, Konishi, Kim, and Liu}]{ke2023cpt}
Zixuan Ke, Yijia Shao, Haowei Lin, Tatsuya Konishi, Gyuhak Kim, and Bing Liu. 2023.
\newblock Continual pre-training of language models.
\newblock \emph{arXiv preprint arXiv:2302.03241}.

\bibitem[{Lei et~al.(2024)Lei, Li, and Chen}]{autocoder}
Bin Lei, Yuchen Li, and Qiuwu Chen. 2024.
\newblock Autocoder: Enhancing code large language model with$\backslash$textsc $\{$AIEV-Instruct$\}$.
\newblock \emph{arXiv preprint arXiv:2405.14906}.

\bibitem[{Li et~al.(2023)Li, Allal, Zi, Muennighoff, Kocetkov, Mou, Marone, Akiki, Li, Chim, Liu, Zheltonozhskii, Zhuo, Wang, Dehaene, Davaadorj, Lamy-Poirier, Monteiro, Shliazhko, Gontier, Meade, Zebaze, Yee, Umapathi, Zhu, Lipkin, Oblokulov, Wang, Murthy, Stillerman, Patel, Abulkhanov, Zocca, Dey, Zhang, Fahmy, Bhattacharyya, Yu, Singh, Luccioni, Villegas, Kunakov, Zhdanov, Romero, Lee, Timor, Ding, Schlesinger, Schoelkopf, Ebert, Dao, Mishra, Gu, Robinson, Anderson, Dolan-Gavitt, Contractor, Reddy, Fried, Bahdanau, Jernite, Ferrandis, Hughes, Wolf, Guha, von Werra, and de~Vries}]{starcoder}
Raymond Li, Loubna~Ben Allal, Yangtian Zi, Niklas Muennighoff, Denis Kocetkov, Chenghao Mou, Marc Marone, Christopher Akiki, Jia Li, Jenny Chim, Qian Liu, Evgenii Zheltonozhskii, Terry~Yue Zhuo, Thomas Wang, Olivier Dehaene, Mishig Davaadorj, Joel Lamy-Poirier, João Monteiro, Oleh Shliazhko, Nicolas Gontier, Nicholas Meade, Armel Zebaze, Ming-Ho Yee, Logesh~Kumar Umapathi, Jian Zhu, Benjamin Lipkin, Muhtasham Oblokulov, Zhiruo Wang, Rudra Murthy, Jason Stillerman, Siva~Sankalp Patel, Dmitry Abulkhanov, Marco Zocca, Manan Dey, Zhihan Zhang, Nour Fahmy, Urvashi Bhattacharyya, Wenhao Yu, Swayam Singh, Sasha Luccioni, Paulo Villegas, Maxim Kunakov, Fedor Zhdanov, Manuel Romero, Tony Lee, Nadav Timor, Jennifer Ding, Claire Schlesinger, Hailey Schoelkopf, Jan Ebert, Tri Dao, Mayank Mishra, Alex Gu, Jennifer Robinson, Carolyn~Jane Anderson, Brendan Dolan-Gavitt, Danish Contractor, Siva Reddy, Daniel Fried, Dzmitry Bahdanau, Yacine Jernite, Carlos~Muñoz Ferrandis, Sean Hughes, Thomas Wolf, Arjun Guha, Leandro von
  Werra, and Harm de~Vries. 2023.
\newblock \href {https://arxiv.org/abs/2305.06161} {Starcoder: may the source be with you!}
\newblock \emph{Preprint}, arXiv:2305.06161.

\bibitem[{Lin et~al.(2024)Lin, Huang, Ye, Chen, Wang, Li, Ma, Wan, Zou, and Liang}]{rectified_sft_scaling_law}
Haowei Lin, Baizhou Huang, Haotian Ye, Qinyu Chen, Zihao Wang, Sujian Li, Jianzhu Ma, Xiaojun Wan, James Zou, and Yitao Liang. 2024.
\newblock Selecting large language model to fine-tune via rectified scaling law.
\newblock \emph{arXiv preprint arXiv:2402.02314}.

\bibitem[{Lu et~al.(2023)Lu, Zhu, Han, Zhao, Mac~Namee, and Tan}]{lu-makes-plm-0-shot}
Jinghui Lu, Dongsheng Zhu, Weidong Han, Rui Zhao, Brian Mac~Namee, and Fei Tan. 2023.
\newblock What makes pre-trained language models better zero-shot learners?
\newblock In \emph{Proceedings of the 61st Annual Meeting of the Association for Computational Linguistics (Volume 1: Long Papers)}, pages 2288--2303.

\bibitem[{Luo et~al.(2023)Luo, Yang, Meng, Li, Zhou, and Zhang}]{empirical_cv_catastrophic_forgetting}
Yun Luo, Zhen Yang, Fandong Meng, Yafu Li, Jie Zhou, and Yue Zhang. 2023.
\newblock An empirical study of catastrophic forgetting in large language models during continual fine-tuning.
\newblock \emph{arXiv preprint arXiv:2308.08747}.

\bibitem[{Pandey(2024)}]{data_scaling_law}
Rohan Pandey. 2024.
\newblock gzip predicts data-dependent scaling laws.
\newblock \emph{arXiv preprint arXiv:2405.16684}.

\bibitem[{Que et~al.(2024)Que, Liu, Zhang, Zhang, Qu, Ma, Duan, Bai, Wang, Zhang et~al.}]{ali_cpt_scaling_law}
Haoran Que, Jiaheng Liu, Ge~Zhang, Chenchen Zhang, Xingwei Qu, Yinghao Ma, Feiyu Duan, Zhiqi Bai, Jiakai Wang, Yuanxing Zhang, et~al. 2024.
\newblock D-cpt law: Domain-specific continual pre-training scaling law for large language models.
\newblock \emph{arXiv preprint arXiv:2406.01375}.

\bibitem[{Rockafellar(1993)}]{lagrange}
R~Tyrrell Rockafellar. 1993.
\newblock Lagrange multipliers and optimality.
\newblock \emph{SIAM review}, 35(2):183--238.

\bibitem[{Touvron et~al.(2023{\natexlab{a}})Touvron, Lavril, Izacard, Martinet, Lachaux, Lacroix, Rozière, Goyal, Hambro, Azhar, Rodriguez, Joulin, Grave, and Lample}]{llama}
Hugo Touvron, Thibaut Lavril, Gautier Izacard, Xavier Martinet, Marie-Anne Lachaux, Timothée Lacroix, Baptiste Rozière, Naman Goyal, Eric Hambro, Faisal Azhar, Aurelien Rodriguez, Armand Joulin, Edouard Grave, and Guillaume Lample. 2023{\natexlab{a}}.
\newblock \href {https://arxiv.org/abs/2302.13971} {Llama: Open and efficient foundation language models}.
\newblock \emph{Preprint}, arXiv:2302.13971.

\bibitem[{Touvron et~al.(2023{\natexlab{b}})Touvron, Martin, Stone, Albert, Almahairi, Babaei, Bashlykov, Batra, Bhargava, Bhosale, Bikel, Blecher, Ferrer, Chen, Cucurull, Esiobu, Fernandes, Fu, Fu, Fuller, Gao, Goswami, Goyal, Hartshorn, Hosseini, Hou, Inan, Kardas, Kerkez, Khabsa, Kloumann, Korenev, Koura, Lachaux, Lavril, Lee, Liskovich, Lu, Mao, Martinet, Mihaylov, Mishra, Molybog, Nie, Poulton, Reizenstein, Rungta, Saladi, Schelten, Silva, Smith, Subramanian, Tan, Tang, Taylor, Williams, Kuan, Xu, Yan, Zarov, Zhang, Fan, Kambadur, Narang, Rodriguez, Stojnic, Edunov, and Scialom}]{llama2}
Hugo Touvron, Louis Martin, Kevin Stone, Peter Albert, Amjad Almahairi, Yasmine Babaei, Nikolay Bashlykov, Soumya Batra, Prajjwal Bhargava, Shruti Bhosale, Dan Bikel, Lukas Blecher, Cristian~Canton Ferrer, Moya Chen, Guillem Cucurull, David Esiobu, Jude Fernandes, Jeremy Fu, Wenyin Fu, Brian Fuller, Cynthia Gao, Vedanuj Goswami, Naman Goyal, Anthony Hartshorn, Saghar Hosseini, Rui Hou, Hakan Inan, Marcin Kardas, Viktor Kerkez, Madian Khabsa, Isabel Kloumann, Artem Korenev, Punit~Singh Koura, Marie-Anne Lachaux, Thibaut Lavril, Jenya Lee, Diana Liskovich, Yinghai Lu, Yuning Mao, Xavier Martinet, Todor Mihaylov, Pushkar Mishra, Igor Molybog, Yixin Nie, Andrew Poulton, Jeremy Reizenstein, Rashi Rungta, Kalyan Saladi, Alan Schelten, Ruan Silva, Eric~Michael Smith, Ranjan Subramanian, Xiaoqing~Ellen Tan, Binh Tang, Ross Taylor, Adina Williams, Jian~Xiang Kuan, Puxin Xu, Zheng Yan, Iliyan Zarov, Yuchen Zhang, Angela Fan, Melanie Kambadur, Sharan Narang, Aurelien Rodriguez, Robert Stojnic, Sergey Edunov, and Thomas
  Scialom. 2023{\natexlab{b}}.
\newblock \href {https://arxiv.org/abs/2307.09288} {Llama 2: Open foundation and fine-tuned chat models}.
\newblock \emph{Preprint}, arXiv:2307.09288.

\bibitem[{Yao and Wang(2023)}]{nanolm}
Yiqun Yao and Yequan Wang. 2023.
\newblock Research without re-search: Maximal update parametrization yields accurate loss prediction across scales.
\newblock \emph{arXiv preprint arXiv:2304.06875}.

\bibitem[{Ye et~al.(2024)Ye, Liu, Sun, Zhou, Zhan, and Qiu}]{pujianglab_mixing_law}
Jiasheng Ye, Peiju Liu, Tianxiang Sun, Yunhua Zhou, Jun Zhan, and Xipeng Qiu. 2024.
\newblock Data mixing laws: Optimizing data mixtures by predicting language modeling performance.
\newblock \emph{arXiv preprint arXiv:2403.16952}.

\bibitem[{Y{\i}ld{\i}z et~al.(2024)Y{\i}ld{\i}z, Ravichandran, Punia, Bethge, and Ermis}]{cpt_llm_survey}
{\c{C}}a{\u{g}}atay Y{\i}ld{\i}z, Nishaanth~Kanna Ravichandran, Prishruit Punia, Matthias Bethge, and Beyza Ermis. 2024.
\newblock Investigating continual pretraining in large language models: Insights and implications.
\newblock \emph{arXiv preprint arXiv:2402.17400}.

\bibitem[{Yuan et~al.(2023)Yuan, Yuan, Li, Dong, Tan, and Zhou}]{scaling_relationship_loss_ali}
Zheng Yuan, Hongyi Yuan, Chengpeng Li, Guanting Dong, Chuanqi Tan, and Chang Zhou. 2023.
\newblock Scaling relationship on learning mathematical reasoning with large language models.
\newblock \emph{arXiv preprint arXiv:2308.01825}.

\bibitem[{Zhang et~al.(2024)Zhang, Wu, Li, Yang, Zhao, Jiang, and Tan}]{zhanghengyuan_balancing}
Hengyuan Zhang, Yanru Wu, Dawei Li, Zacc Yang, Rui Zhao, Yong Jiang, and Fei Tan. 2024.
\newblock Balancing speciality and versatility: a coarse to fine framework for supervised fine-tuning large language model.
\newblock \emph{arXiv preprint arXiv:2404.10306}.

\end{thebibliography}

\appendix
\onecolumn

\section{LLM Configurations}
The detailed parameters of the LLM configurations are listed in Table \ref{configuration_llms}.

\begin{table*}[h!]
\centering
\caption{Configurations of the LLMs.}\label{configuration_llms}
\begin{tabular}{lcccc}
\toprule
Model Size & 460\textrm{m} & 940M & 1.6B & 3.1B \\
\midrule
hidden size & 1024 & 1536 & 2048 & 2560 \\
intermediate size & 3072 & 4608 & 6144 & 7680 \\
number of attention heads & 32 & 32 & 32 & 32 \\
number of layers & 24 & 24 & 24 & 32 \\
vocabulary size & 65632& 65632& 65632& 65632 \\
\bottomrule
\end{tabular}
\end{table*}

\section{Mathematical derivation}
\subsection{Notation}
\begin{itemize}
    \item \( S \) - represents the model sizes.
    \item \( M \) - the pre-trained large language model.
    \item \( \mathcal{D}_{\text{gen}} \) - the general dataset.
    \item \( \mathcal{D}_{\text{dom}} \) - the domain-specific dataset.
    \item \( R \) - the mixture ratio of the domain-specific data.
    \item \( \mathcal{D}_{R} \) - the total mixed dataset with \( R\% \) domain-specific data.
    \item \( \epsilon \) - tolerance for the general loss increase.
    \item \( \mathcal{L}_{\text{gen}}^{\text{CPT}} \) - the general loss.
    \item \( \mathcal{L}_{\text{dom}}^{\text{CPT}} \) - the domain-specific loss.
    \item \( \mathcal{L}_{\Delta\text{gen}} \) - the increment in general loss.
    \item \( \mathcal{L}_{\Delta\text{dom}} \) - the increment in domain-specific loss.
    \item \( F \) - the loss function of CPT expressed as the Lagrangian.
    \item \( T \) - the amount of training tokens (related to the number of iterations, training steps, or the total volume of training data).
    \item \( \lambda \) - the Lagrange multiplier used to enforce the constraint on general loss while minimizing domain-specific loss.
    \item \( T_{\text{max}} \) - the maximum training tokens for CPT.
    \item \( T_0 \) - a point on the training curve where, after training at \( T_0 \) and continuing the training, the feasible mixture ratio is observed.
    \item \( \mathbb{A} \) - the set of mixture ratios that satisfying CPT objective.
    \item \( \mathbb{F} \) - the set of Feasible Mixture Ratios (feasible mixture ratio).
    \item \( R_{\text{CMR}} \) - the Critical Mixture Ratio (CMR), which is the optimal mixture ratio that minimizes the loss function within the feasible set.
    \item \(\alpha_1\), \(\alpha_2\), \(\alpha_3\) - parameters to be fitted representing coefficients in the power-law functions for the increment of loss.
    \item \(\beta_1\), \(\beta_2\) - parameters to be fitted representing constants in the power-law functions for the increment of loss.
    \item \( s_1 \), \( s_2 \), \( s_3 \) - parameters to be fitted representing the exponents in the power-law functions for the increment of loss.
\end{itemize}

\subsection{Feasible mixture ratio}\label{apd:math}
Given that \(N\) is fixed, the 0bjective of CPT in Equation~\ref{eq:opt_function} can be transformed into:
\begin{equation}
\begin{aligned}
F(R, T, \lambda) = &(\mathcal{L}_{\text{dom}}(M_{S}) +  \mathcal{L}_{\Delta\text{dom}}({R}, T)) + \lambda ( \mathcal{L}_{\Delta\text{gen}}(\mathcal{R}, T)- \epsilon ),
\end{aligned}
\label{eq:opt_function_delta}
\end{equation}
where $\mathcal{L}_{\text{dom}}(\text{CPT}(M_{S}; \mathcal{D}_{R}, T))$ is split into the value at $T=0$, $\mathcal{L}_{\text{dom}}(M_{S})$ and the increment $\mathcal{L}_{\Delta\text{dom}}$. The corresponding 
\begin{equation}
\begin{aligned}
R^* = \text{argmin}_{R} {F}(R, T, \lambda) \\
\text{s.t.} \quad 
\begin{cases} 
\mathcal{L}_{\Delta \text{gen}}(R, T) \leq \epsilon \\
R \geq 0 \\
T_{\max} \geq T \geq 0 \\
\lambda \geq 0.
\end{cases}
\end{aligned}
\end{equation}
For a given mixture ratio $R$, if the training progresses ($T$ increases), and the objective function~(Equation~\ref{eq:opt_function_delta}) shows a decreasing trend, it indicates that the current proportion can lead to the continuation of training towards the expected goal. The trend of the objective function $F$ increasing with training can be reflected by its partial derivative with respect to $T$ : 
\begin{equation}
\begin{aligned}
& \left. \frac{\partial F(R, T, \lambda)}{\partial T} \right|_{R, \lambda} = \left. \frac{\partial \left(\mathcal{L}_{\text{dom}}(M_{S}) + \mathcal{L}_{\Delta\text{dom}}(R, T)\right)}{\partial T} \right|_{R,\lambda} \quad + \lambda \left. \frac{\partial \left(\mathcal{L}_{\Delta \text{gen}}(R, T) - \epsilon\right)}{\partial T} \right|_{R,\lambda}.
\end{aligned}
\label{eq:partial_derivative}
\end{equation}
Since \(\mathcal{L}_{\text{dom}}(M_{S})\) and \(-\lambda \epsilon\) are constants with respect to \( T \), their derivatives are zero. Thus, we simplify to:
\begin{equation}
\begin{aligned}
\left. \frac{\partial F(R, T, \lambda)}{\partial T} \right|_{R,\lambda}  \left. \frac{\partial \mathcal{L}_{\Delta \text{dom}}(R, T)}{\partial T} \right|_{R,\lambda}  + \lambda \left. \frac{\partial \mathcal{L}_{\Delta \text{gen}}(R, T)}{\partial T} \right|_{R,\lambda}.
\end{aligned}
\label{eq:partial_derivative_simplified}
\end{equation}
If the training objective under the fixed ratio progresses as expected, there should be at least one point during the training process (\(0 \leq T \leq T_{\text{max}}\)) where this partial derivative is less than or equal to 0. From this, we can define a feasible proportion curve that should satisfy the following inequality conditions: 
\begin{equation}
\exists \, T \in [0, T_{\text{max}}] \, : \, \left. \frac{\partial F(R, T, \lambda)}{\partial T} \right|_{R,\lambda} \leq 0
\label{eq:derived_fangchen}
\end{equation}
This means that we only need to determine whether the solution $T$ of the above inequality~\ref{eq:derived_fangchen} belongs to $[0, T_{\mathrm{max}}]$ in order to judge whether the current training meets the target. Setting it equal to zero to figure out:
\begin{equation}
\left. \frac{\partial F(R, T, \lambda)}{\partial T} \right|_{R, \lambda} = 0
\label{eq:partial_derivative_zero}
\end{equation}
Setting the equation to zero and further simplifying to express it :
\begin{equation}
\begin{aligned}
\left. \frac{\partial \mathcal{L}_{\Delta\text{dom}}(R, T)}{\partial T} \right|_{R} 
+ \lambda \left. \frac{\partial \mathcal{L}_{\Delta\text{gen}}(R, T)}{\partial T} \right|_{R} = 0
\end{aligned}
\label{eq:partial_derivative_plus_zero}
\end{equation}
To derive the following equation using the chain rule:
\begin{equation}
\left. \frac{\partial \mathcal{L}_{\Delta\text{dom}}(R, T)}{\partial T} \right|_{R} = -\lambda \left. \frac{\partial \mathcal{L}_{\Delta\text{gen}}(R, T)}{\partial T} \right|_{R}
\end{equation}
By isolating \(\left. \frac{\partial \mathcal{L}_{\Delta\text{dom}}(R, T)}{\partial T} \right|_{R}\), we get:
\begin{equation}
\left. \frac{\partial \mathcal{L}_{\Delta\text{dom}}(R, T)}{\partial T} \right|_{R} = -\lambda \left. \frac{\partial \mathcal{L}_{\Delta\text{gen}}(R, T)}{\partial T} \right|_{R}
\end{equation}
Using the chain rule, we have:
\begin{equation}
\begin{aligned}
\left. \frac{\partial \mathcal{L}_{\Delta\text{gen}}(R, T)}{\partial T} \right|_{R} &= \left. \frac{\partial \mathcal{L}_{\Delta\text{gen}}(R, T)}{\partial \mathcal{L}_{\Delta\text{dom}}(R, T)} \right|_{R} \quad \cdot \left. \frac{\partial \mathcal{L}_{\Delta\text{dom}}(R, T)}{\partial T} \right|_{R}
\end{aligned}
\end{equation}
By substituting this into the given equation, we get:
\begin{equation}
\begin{aligned}
 \left. \frac{\partial \mathcal{L}_{\Delta\text{dom}}(R, T)}{\partial T} \right|_{R}  = -\lambda \left( \frac{\partial \mathcal{L}_{\Delta\text{gen}}(R, T)}{\partial \mathcal{L}_{\Delta\text{dom}}(R, T)} \cdot \left. \frac{\partial \mathcal{L}_{\Delta\text{dom}}(R, T)}{\partial T} \right|_{R} \right)
\end{aligned}
\end{equation}
Assuming \(\left. \frac{\partial \mathcal{L}_{\Delta\text{dom}}(R, T)}{\partial T} \right|_{R} \neq 0\), we can cancel the terms:
\begin{equation}
1 = -\lambda \cdot \left. \frac{\partial \mathcal{L}_{\Delta\text{gen}}(R, T)}{\partial \mathcal{L}_{\Delta\text{dom}}(R, T)} \right|_{R}
\end{equation}
Thus, we obtain:
\begin{equation}
\left. \frac{\partial \mathcal{L}_{\Delta\text{gen}}(R, T)}{\partial \mathcal{L}_{\Delta\text{dom}}(R, T)} \right|_{R} = -\frac{1}{\lambda}
\label{eq:lambda_1}
\end{equation}
Since \(\lambda > 0\), the above derivative is a negative number. For a specific \(R\), if there exist points on the training curve where the partial derivatives of the two \(\Delta\) values are equal to \(\frac{1}{\lambda}\), then the ratio is consistent with the expected goal of continual pretraining. These ratios are called feasible mixture ratios, and their set is denoted as \(\mathbb{F}\). This is consistent with the feasible mixture ratios marked in Figure~\ref{fig:3d-model}.

\subsection{Fitting}

Following the previous work~\cite{openai_scaling_law,chinchilla_scaling_law}, we have adopted the power-law as the parametric forms, which is different from other mixture law study~\cite{pujianglab_mixing_law}. Previous work has shown that the model parameter $N$ and the amount of data training $T$ are independently related to the power law of loss. However, one point that our work related to power law is different. First, the function we choose to fit is the increment of Loss. Second, due to the phenomenon of general loss increasing first and then decreasing, in order to better fit the data, we used a two-term power-law function. According to Equation~\ref{eq:opt_function_delta}, the data mixture scaling law for CPT training is defined as follows: 

Given:
\begin{equation}
\left\{
\begin{aligned}
\mathcal{L}_{\Delta \text{dom}}(T) &= \alpha_1 \cdot T^{s1} + \beta_1 ,\\
\mathcal{L}_{\Delta \text{gen}}(T) &= \alpha_2 \cdot T^{s2} + \alpha_3 \cdot T^{s3} + \beta_2.
\end{aligned}
\right.
\end{equation}
where \(\alpha_1\), \(\alpha_2\), \(\alpha_3\), \(\beta_1\), \(\beta_2\), \(s1\), \(s2\), and \(s3\) are parameters to be fitted.

First, according to the definition of feasible mixture ratios, we can solve feasible mixture ratios under the setting of data mixture scaling law. As the fitting at this time is an extrapolation of the training quantity, R is a fixed value. For simplicity, we no longer explicitly write R, so both $\mathcal{L}_{\Delta \text{dom}}$ and $\mathcal{L}_{\Delta \text{gen}}$ are univariate functions of $T$.
First, differentiate \( \mathcal{L}_{\Delta \text{dom}}(T)\) with respect to \(T\):
\begin{equation}
\begin{aligned}
\frac{d}{dT} \mathcal{L}_{\Delta \text{dom}}(T) &= \frac{d}{dT} (\alpha_1 \cdot T^{s1} + \beta_1) \\
&= \alpha_1 \cdot s1 \cdot T^{s1 - 1}.
\label{eq:para_d_l_dom}
\end{aligned}
\end{equation}
Next, differentiate \( \mathcal{L}_{\Delta\text{gen}}(T)\) with respect to \(T\):
\begin{equation}
\begin{aligned}
\frac{d}{dT}  \mathcal{L}_{\Delta\text{gen}}(T) = \frac{d}{dT} (\alpha_2 \cdot T^{s2} + \alpha_3 \cdot T^{s3} + \beta_2) = \alpha_2 \cdot s2 \cdot T^{s2 - 1} + \alpha_3 \cdot s3 \cdot T^{s3 - 1}.
\label{eq:para_d_l_gen}
\end{aligned}
\end{equation}
According to the the expected CPT trend in Equation~\ref{eq:derived_fangchen}, we need to figure whether the critical $T_0$ that meets this condition is in the effective range $[0,T_{\mathrm{max}}]$. Therefore, the solution for the Equation~\ref{eq:partial_derivative_plus_zero} is important, which can be solved as Equation~\ref{eq:partial_derivative_plus_zero}:
\begin{equation}
\begin{aligned}
\frac{d}{dT} \mathcal{L}_{\Delta \text{dom}}(T)  + \lambda \frac{d}{dT}  \mathcal{L}_{\Delta\text{gen}}(T) =0
\end{aligned}
\end{equation}
Substitute Equation~\ref{eq:para_d_l_dom} and Equation~\ref{eq:para_d_l_gen} respectively, we get:
\begin{equation}
\begin{aligned}
\alpha_1 \cdot s1 \cdot T^{s1 - 1} &+ \lambda (\alpha_2 \cdot s2 \cdot T^{s2 - 1} + \alpha_3 \cdot s3 \cdot T^{s3 - 1}) = 0
\end{aligned}
\end{equation}
Further simplifying:
\begin{equation}
\begin{aligned}
\alpha_1 \cdot s1 \cdot T^{s1 - 1} &+ \lambda \alpha_2 \cdot s2 \cdot T^{s2 - 1} + \lambda \alpha_3 \cdot s3 \cdot T^{s3 - 1} = 0
\end{aligned}
\end{equation}
To solve for \(T\), we can factor out \(T\) terms:
\begin{equation}
\begin{aligned}
T^{s1 - 1} \bigl( & \alpha_1 \cdot s1  + \lambda \alpha_2 \cdot s2 \cdot T^{s2 - s1}  + \lambda \alpha_3 \cdot s3 \cdot T^{s3 - s1} \bigr) = 0
\end{aligned}
\end{equation}
Therefore, the critical points \(T_0\) can be solved by:
\begin{equation}
\begin{aligned}
T_0^{s2 - s1} = -\frac{\alpha_1 \cdot s1}{\lambda \alpha_2 \cdot s2} \quad - \frac{\lambda \alpha_3 \cdot s3 \cdot T_0^{s3 - s1}}{\lambda \alpha_2 \cdot s2}
\end{aligned}
\end{equation}
Solving for \(T_0\):
\begin{equation}
\begin{aligned}
T_0^{s2 - s1} &= -\frac{\alpha_1 \cdot s1 + \lambda \alpha_3 \cdot s3 \cdot T_0^{s3 - s1}}{\lambda \alpha_2 \cdot s2}
\end{aligned}
\end{equation}
\begin{equation}
\begin{aligned}
T_0^{s2 - s1} = -\frac{\alpha_1 \cdot s1}{\lambda \alpha_2 \cdot s2} - \frac{\lambda \alpha_3 \cdot s3 \cdot T_0^{s3 - s1}}{\lambda \alpha_2 \cdot s2}
\end{aligned}
\end{equation}
Thus, the solution for \(T_0\) in terms of the original parameters is:
\begin{equation}
\begin{aligned}
T_0 = \left[-\frac{\alpha_1 \cdot s1}{\lambda \alpha_2 \cdot s2} \left(1 + \frac{\alpha_3 \cdot s3}{\alpha_2 \cdot s2} T_0^{s3 - s2}\right)^{-1}\right]^{\frac{1}{s2 - s1}}
\end{aligned}
\end{equation}

\section{Justification of Tolerance Value}

\subsection{How was this determined?}
Tolerance $\epsilon$ in Equation~\ref{eq:target1} depends on the importance of maintaining general abilities for CPT goals (Definition~\ref{def:CPT object}). We set $\epsilon = 0.05$ based on empirical results by observation. However, \textbf{the value of $\epsilon$ does not affect the conclusions and analysis} presented in this paper. In practical applications, its setting is related to the researcher's considerations of relevant factors (application scenarios, resource situation, etc.) for CPT.

As shown in Figure~\ref{fig:predicting_T} and Figure~\ref{fig:arxiv_T_prediction}, $\Delta_\text{General loss}$ typically peaks at values much smaller than 0.05. This is because these figures depict relatively small mixture ratios ($\mathcal{R} \leq \frac{1}{2}$), where general losses initially increase before decreasing. As the mixture ratio grows, $\Delta_\text{General Loss}$ continues to rise and eventually exceeds 0.05 ($\mathcal{R} \to 1$). This trend can be initially observed in Figure~\ref{fig:3d-model} and Figure~\ref{fig:3d-all}.

\subsection{How does this value reflect the constraint in practice?}

    As detailed in Definition~\ref{def:FMR}, the tolerance value $\epsilon$ is used to determine a range of mixture ratios $\mathbb{A}$ that do not lead to excessive increases in general losses (i.e., unaffordable losses in general capabilities). \textbf{In practice, we use the tolerance to identify the upper bound of feasible mixture ratios}. For example, an excessive mixture ratio may cause the general loss to exceed the tolerance threshold. Therefore, even if the general loss under this ratio reaches a plateau or decreases, it is still considered infeasible.

\section{Table}~\label{more_table}
To illustrate the accuracy of our fitting, we provide the relative error for the fitted curves in Table~\ref{tab:domain_ratio_predict}, along with the MSE and $R^2$ values for the power-law fitting of $\Delta$General ($\Delta$Domain) loss as a function of training tokens $T$ in Table~\ref{tab:MSE_R2}. For better reproducibility, the fitted scaling law coefficients are presented in Table~\ref{tab:coef_R_T} and Table~\ref{tab:coef_T0}, which correspond to \S~\ref{subsec:predicting_cmr}. In summary, we found that the power function accurately fits the loss with respect to model size, data mixture ratio, and token volume, using the \texttt{Scipy} library to estimate the function's coefficients.

\begin{table}[h!]
  \centering
  \renewcommand{\arraystretch}{1.3}
  \setlength{\tabcolsep}{5pt} 
  \begin{tabular}{>{\raggedright}p{2cm}cccc}
    \hline
    \textbf{Ratio} & \textbf{460\textbf{M}} & \textbf{940M} & \textbf{1.6B} & \textbf{3.1B} \\
    \hline
    100\% & 1.4628 & 1.3723 & 1.3242 & 1.2585 \\
    75\%  & 1.4844 & 1.3910 & 1.3416 & 1.2750 \\
    50\%  & 1.5122 & 1.4155 & 1.3643 & 1.2965 \\
    33\%  & 1.5387 & 1.4385 & 1.3854 & 1.3170 \\
    25\%-gt  & 1.5561 & 1.4538 & 1.3994 & 1.3305 \\
    25\%-pred & 1.5566 & 1.4546 & 1.3999 & 1.3303 \\
    \textbf{Difference} & \textbf{0.03\%} & \textbf{0.05\%} & \textbf{0.03\%} & \textbf{0.02\%} \\
    \hline
  \end{tabular}
  \caption{Domain Proportion and Predicted/Actual Value Relative Error}
  \label{tab:domain_ratio_predict}
  \vspace{-6pt} 
\end{table}

\begin{table*}[h!]
  \centering
  \small
  \renewcommand{\arraystretch}{1.3}
  \setlength{\tabcolsep}{1pt} 
  \begin{tabular}{cl|cccc|cccc}
    \hline
    \multirow{2}{*}{\textbf{Metric}} & \multirow{2}{*}{\textbf{Ratio}} & \multicolumn{4}{c}{\textbf{General}} & \multicolumn{4}{c}{\textbf{Domain}} \\
    && \textbf{460\textbf{M}} & \textbf{940M} & \textbf{1.6B} & \textbf{3.1B} & \textbf{460\textbf{M}} & \textbf{940M} & \textbf{1.6B} & \textbf{3.1B} \\
    \hline
    \multirow{6}{*}{\textbf{MSE}} & 100\% & 1.9394e-10 & 2.5695e-10 & 9.8058e-10 & 2.2880e-11 & 9.7830e-08 & 7.6174e-08 & 6.4577e-08 & 4.4057e-08 \\
    & 75\%  & 5.2270e-11 & 7.7104e-15 & 3.4402e-12 & 1.5432e-11 & 1.2283e-07 & 7.6940e-08 & 7.1749e-08 & 4.4160e-08 \\
    & 50\%  & 1.5340e-10 & 4.4162e-11 & 1.5992e-09 & 1.6405e-10 & 1.2539e-07 & 7.0535e-08 & 5.1893e-08 & 3.8559e-08 \\
    & 33.3\%  & 5.2538e-11 & 1.1070e-10 & 5.5883e-11 & 5.4041e-11 & 1.1904e-07 & 6.9371e-08 & 5.8162e-08 & 4.4630e-08 \\
    & 25\%  & 7.3045e-11 & 4.7677e-11 & 8.7140e-11 & 1.6598e-14 & 1.0966e-07 & 7.0327e-08 & 4.7272e-08 & 4.6702e-08 \\
    & 12.5\% & 6.9011e-11 & 8.9891e-11 & 7.1858e-11 & 9.2656e-11 & 8.1609e-08 & 7.2597e-08 & 5.1854e-08 & 4.4091e-08 \\
    \hline
    \multirow{6}{*}{\textbf{$R^2$}} & 100\%  & 0.9999 & 0.9999 & 0.9998 & 0.9989 & 0.9957 & 0.9969 & 0.9969 & 0.9978 \\
    & 75\%  & 0.9993 & 0.9999 & 0.9990 & 0.9963 & 0.9954 & 0.9966 & 0.9966 & 0.9975 \\
    & 50\%  & 0.9973 & 0.9966 & 0.9946 & 0.9593 & 0.9951 & 0.9963 & 0.9965 & 0.9971 \\
    & 33.3\%  & 0.9928 & 0.9877 & 0.9818 & 0.9251 & 0.9954 & 0.9959 & 0.9965 & 0.9967 \\
    & 25\%  & 0.9872 & 0.9763 & 0.9659 & 0.8741 & 0.9956 & 0.9959 & 0.9966 & 0.9966 \\
    & 12.5\% & 0.9590 & 0.9438 & 0.9520 & 0.8972 & 0.9974 & 0.9962 & 0.9963 & 0.9965 \\
    \hline
  \end{tabular}
  \caption{The MSE and $R^2$ of the fitting power-law of $\Delta$General ($\Delta$Domain) loss by training tokens $T$}
  \label{tab:MSE_R2}
  \vspace{-6pt} 
\end{table*}

\begin{table*}
  \centering
  \small
  \renewcommand{\arraystretch}{1.3}
  \setlength{\tabcolsep}{1pt} 
  \begin{tabular}{c|c|ccccc|ccc}
    \hline
    \multirow{2}{*}{\textbf{Model Size}} & 
    \multirow{2}{*}{\textbf{Ratio}} & 
    \multicolumn{5}{c}{\textbf{General}} & \multicolumn{3}{c}{\textbf{Domain}} \\
    && \textbf{$\alpha_1$} & \textbf{$s_1$} & \textbf{$\alpha_2$} & \textbf{$s_2$} & \textbf{$\beta_1$} & \textbf{$\alpha_3$} & \textbf{$s_3$} & \textbf{$\beta_2$} \\
    \hline
    \multirow{6}{*}{\textbf{460M}} & 100\% & -0.01502 & 0.17543 & 0.02116 & 0.46219 & 0.00472 & -2.43854 & 0.02229 & 2.24461 \\
    & 75\%  & -0.14742 & 0.64329 & 0.15135 & 0.63990 & 0.00144 & 257.77904 & -0.00022 & -257.95781 \\
    & 50\%  & 0.12446 & 0.57575 & -0.12123 & 0.57941 & 0.00050 & 227.06960 & -0.00023 & -227.23919 \\
    & 33.3\%  & 0.11761 & 0.53594 & -0.11471 & 0.53971 & 0.00002 & 97.25206 & -0.00050 & -97.40981 \\
    & 25\%  & 0.14030 & 0.51526 & -0.13758 & 0.51836 & -0.00018 & 24.62425 & -0.00190 & -24.77263 \\
    & 12.5\%  & 0.13055 & 0.48615 & -0.12803 & 0.48955 & -0.00055 & 0.71989 & -0.08343 & -0.81231 \\
    \hline
    \multirow{6}{*}{\textbf{940M}} & 100\% & 0.00987 & 0.51496 & -0.00521 & 0.00000 & 0.00423 & 262.61092 & -0.00021 & -262.90741 \\
    & 75\%  & 0.00653 & 0.30806 & -0.00500 & 0.00000 & 0.00232 & 259.14017 & -0.00020 & -259.43163 \\
    & 50\%  & 0.09454 & 0.57282 & -0.09209 & 0.57654 & 0.00101 & 258.06226 & -0.00019 & -258.34505 \\
    & 33.3\%  & 0.10418 & 0.52619 & -0.10186 & 0.52970 & 0.00063 & 248.51191 & -0.00018 & -248.78323 \\
    & 25\%  & 0.11006 & 0.51700 & -0.10785 & 0.52040 & 0.00046 & 222.52376 & -0.00020 & -222.78786 \\
    & 12.5\%  & 0.12685 & 0.48822 & -0.12468 & 0.49141 & 0.00013 & 129.70296 & -0.00031 & -129.94852 \\
    \hline
    \multirow{6}{*}{\textbf{1.6B}} & 100\% & 0.00752 & 0.50645 & -0.00385 & 0.00000 & 0.00384 & -0.39825 & 0.09162 & 0.20416 \\
    & 75\%  & 0.06381 & 0.63024 & -0.06167 & 0.63444 & 0.00161 & 210.68118 & -0.00023 & -210.84391 \\
    & 50\%  & 0.07899 & 0.57219 & -0.07702 & 0.57587 & 0.00084 & 258.78683 & -0.00017 & -258.94316 \\
    & 33.3\%  & 0.08952 & 0.54505 & -0.08764 & 0.54858 & 0.00050 & 133.08715 & -0.00032 & -133.23264 \\
    & 25\%  & 0.08747 & 0.53775 & -0.08564 & 0.54152 & 0.00034 & 4.92547 & -0.00864 & -5.06034 \\
    & 12.5\%  & 0.10378 & 0.51241 & -0.10198 & 0.51585 & 0.00007 & 210.06851 & -0.00018 & -210.18657 \\
    \hline
    \multirow{6}{*}{\textbf{3.1B}} & 100\% & 0.00978 & 0.38822 & 0.00000 & 3.42674 & 0.01398 & -0.30116 & 0.10974 & -0.03868 \\
    & 75\%  & 0.04886 & 0.56155 & -0.04660 & 0.56711 & 0.01184 & 265.17268 & -0.00018 & -265.47797 \\
    & 50\%  & 0.06324 & 0.52050 & -0.06100 & 0.52592 & 0.01093 & 156.20051 & -0.00027 & -156.50271 \\
    & 33.3\%  & 0.08406 & 0.51263 & -0.08187 & 0.51724 & 0.01056 & 128.57034 & -0.00031 & -128.86546 \\
    & 25\%  & 0.08084 & 0.51463 & -0.07866 & 0.51973 & 0.01034 & 133.03230 & -0.00028 & -133.32244 \\
    & 12.5\%  & 0.09549 & 0.49146 & -0.09320 & 0.49629 & 0.01001 & 7.55302 & -0.00463 & -7.82759 \\
    \hline
    \end{tabular}
    \caption{Fitting power-law coefficients for different model sizes and the mixture ratios of $\Delta$General and $\Delta$Domain losses as a function of training tokens $T$}
    \label{tab:coef_R_T}
\end{table*}

\begin{table*}
  \centering
  \renewcommand{\arraystretch}{1.3}
  \setlength{\tabcolsep}{1pt} 
  \begin{tabular}{c|ccc}
    \hline
    \textbf{Model Size} & \textbf{$\alpha_4$} & \textbf{$s_4$} & \textbf{$\beta_3$} \\
    \hline
    \textbf{460M} & 0.22524761 & 0.26944345 & -0.48139982 \\
    \textbf{940M} & 0.7520627 & 0.13720245 & -1.06581937 \\
    \textbf{1.6B} & -2.36384831 & -0.15125569 & 1.59223649 \\
    \textbf{3.1B} & -2.5368197 & -0.42071423 & 0.84375368 \\
    \hline
  \end{tabular}
  \caption{Fitted CPT scaling law coefficients for different model sizes in \textit{Finance}.}
  \label{tab:coef_T0}
\end{table*}

\section{Figure}\label{apd:more_figure}
\begin{figure}[t]
  \centering
  \includegraphics[width=\columnwidth]{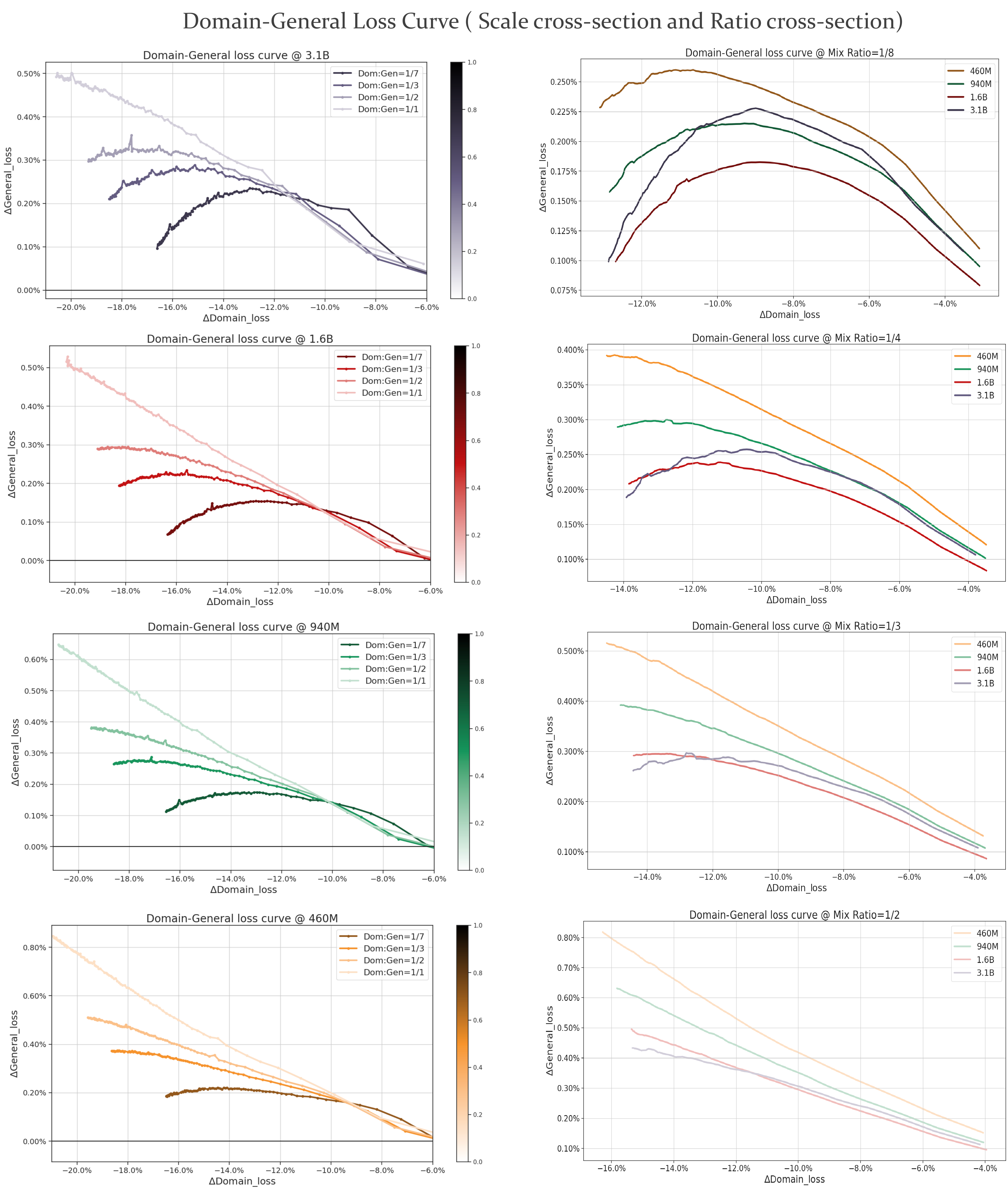}
  \caption{Each cluster represents a different mixing ratio, which is {1/8, 1/4, 1/3, 1/2}. Pay attention to the third set of lines, that is, clusters with a proportion of 1/3. The cross-section of this set of lines is shown on the right.}
  \label{fig:3d-all}
\end{figure}

\begin{figure*}[t]
  \centering
  \includegraphics[width=\columnwidth]{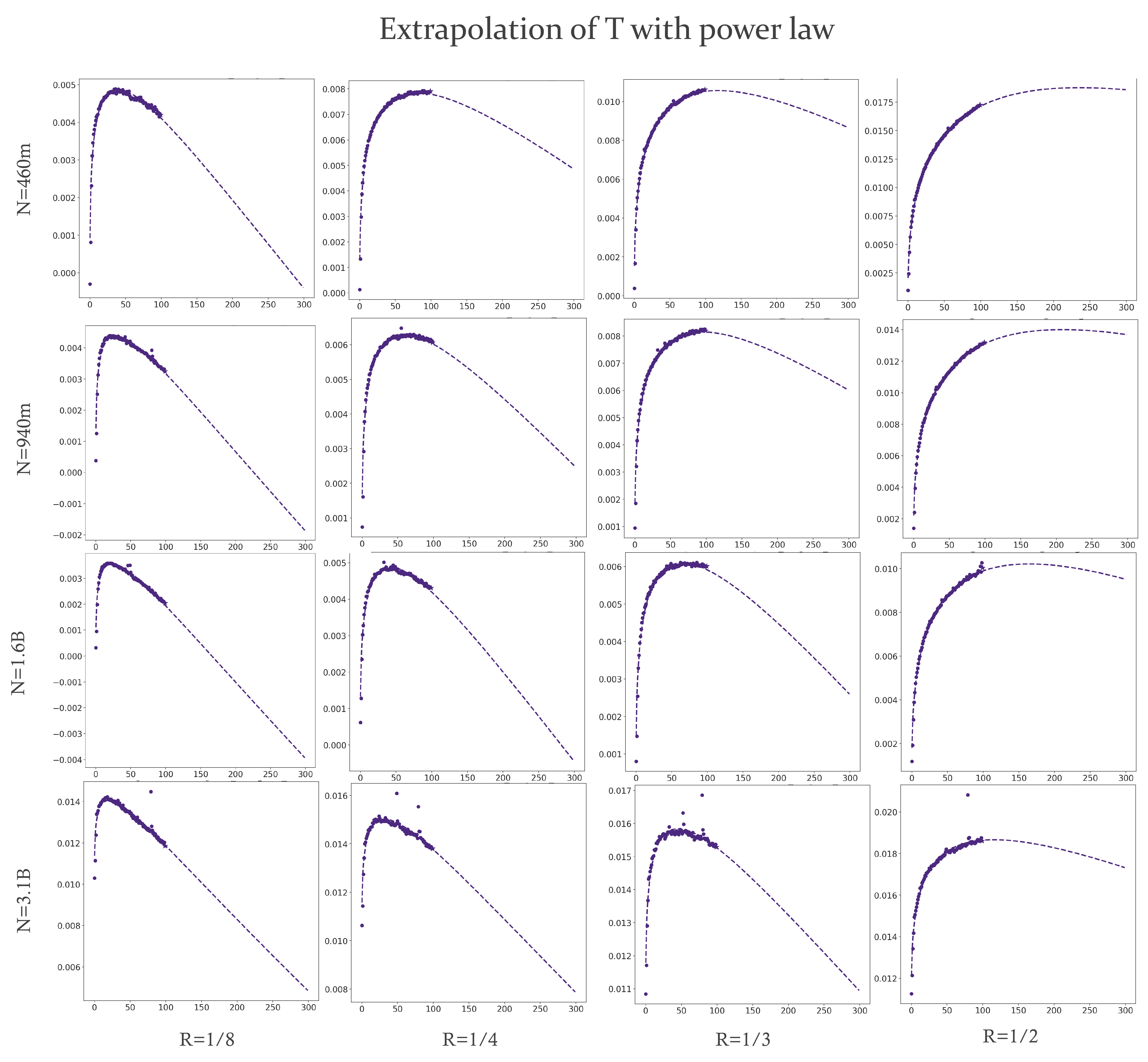}
  \caption{Power laws of training token volume for different model sizes in \textit{Finance}. Compared with the extrapolation of the training volume of the model of the same size to continue training in the \textit{Academic papers} field in Figure~\ref{fig:arxiv_T_prediction}, it can be seen that under the same proportion, the amount of training volume of CPT of Academic Papers is larger where the inflection point appears.}
  \label{fig:extrapolation_of_T}
\end{figure*}

\begin{figure}[t]
  \centering
  \includegraphics[width=\columnwidth]{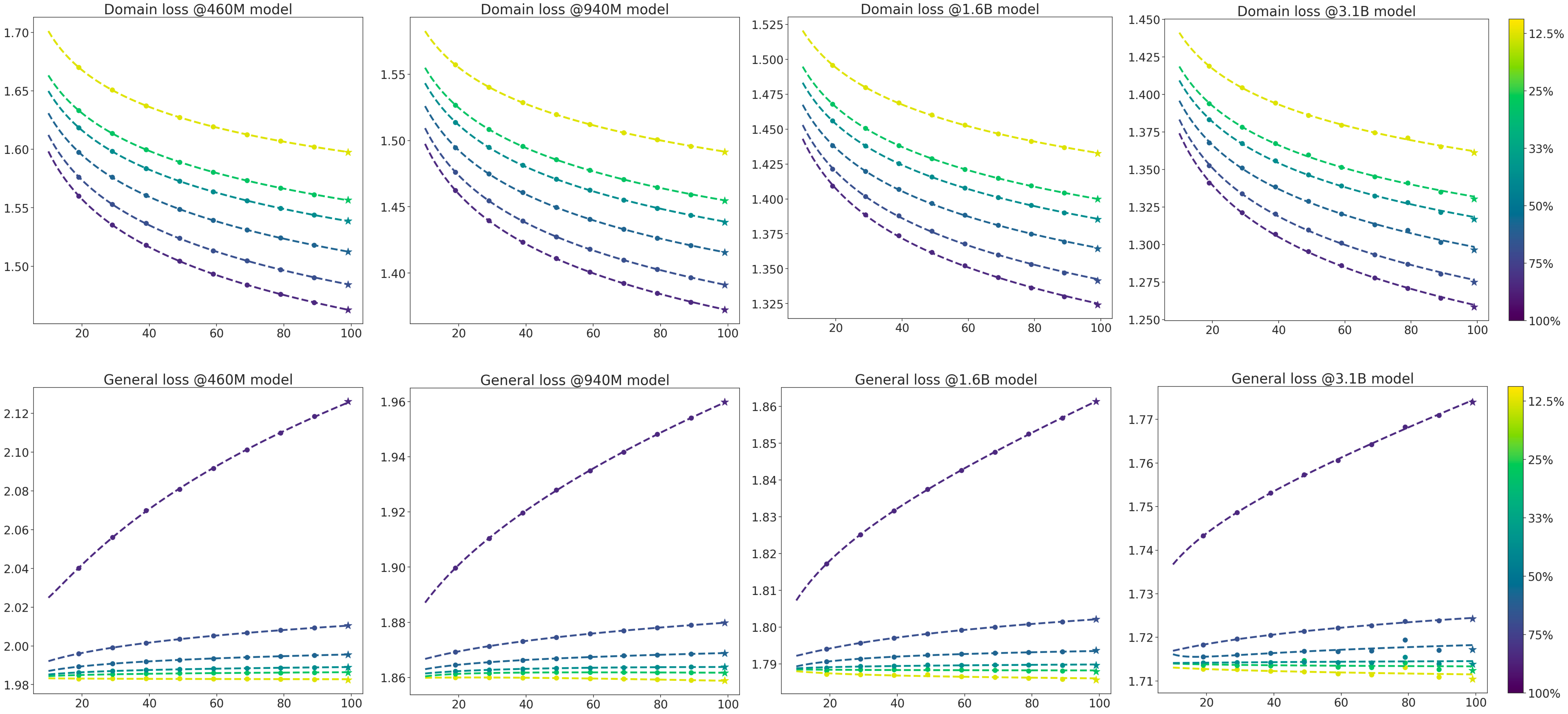}
  \caption{The temperature bar represents the mixture ratio $R$, which takes six values ranging from 1/8 to 1. Different subgraphs are fitting curves that change with the increase of $T$ in the training process for different $M_N$ domain loss and general loss. Overall, the domain loss keeps decreasing during the training process while the general loss keeps increasing. It is worth noting that although the general loss is increasing, the magnitude of its increase is actually very small, especially when the mixture ratio is not very big~($R = \{1/8,1/4, 1/3, 1/2, 3/4\}$), with a total increase of less than $0.02$. The solid circles~($\bullet$) represent real losses, and the stars~(\small \ding{72}) represents the predicted losses. We can see that whether it is general loss or domain loss, the predicted values fall on the fitted curves accurately.}
  \label{fig:token-prediction}
\end{figure}

To comprehensively illustrate the patterns observed in our experiments and our findings, we present the evolution of loss across different mixture ratios and model sizes throughout training in Figure~\ref{fig:3d-all}. Additionally, Figure~\ref{fig:extrapolation_of_T} shows the extrapolation of $T$ using a power-law fit, highlighting the variations in the rise-then-fall trend of general loss under different mixture ratios and model sizes. Finally, the changes and predictions of general loss and domain loss with respect to $T$ under different mixture ratios are illustrated in Figure~\ref{fig:token-prediction}. This effectively explains why we use different forms of power laws to predict $T$ in \S~\ref{subsec:loss_and_token}.

\end{document}